\documentclass[letterpaper, 10 pt, conference]{ieeeconf}
\IEEEoverridecommandlockouts
\usepackage{cite}
\usepackage{amsmath,amssymb,amsfonts}
\usepackage{algorithmic}
\usepackage{graphicx}
\usepackage{textcomp}
\usepackage{mathtools}
\usepackage{comment}
\usepackage[colorinlistoftodos]{todonotes}
\usepackage{tabularx}
\usepackage{array}
\usepackage{booktabs}
\usepackage{caption}
\usepackage{subcaption}
\usepackage{multirow}
\let\labelindent\relax
\usepackage{enumitem}
\usepackage{hyperref}

\newcolumntype{P}[1]{>{\centering\arraybackslash}p{#1}}
\newcolumntype{C}[1]{>{\arraybackslash}p{#1}}

\def\BibTeX{{\rm B\kern-.05em{\sc i\kern-.025em b}\kern-.08em
    T\kern-.1667em\lower.7ex\hbox{E}\kern-.125emX}}
\begin{document}
\bstctlcite{IEEEexample:BSTcontrol}
% \title{\LARGE \bf Task Generalization in Surgical Gesture and Motion Primitive Recognition Models}
%\title{\LARGE \bf Analysis of Granularity Level and Task Generalization in \\Surgical Activity Recognition Models}

% \title{\LARGE \bf Evaluating the Effect of Label Granularity on the \\ Task Generalization of Surgical Activity Recognition}

%\title{\LARGE \bf \textcolor{black}{Evaluating the Effect of Label Granularity on the \\ Task Generalization of Surgical Activity Recognition: \\ A Case Study of TCN using Kinematic Data}}
\title{\LARGE \textcolor{black}{Evaluating the Task Generalization of Temporal Convolutional Networks for Surgical Gesture and Motion Recognition using Kinematic Data}}

\author{Kay Hutchinson$^{*1}$, Ian Reyes$^{2}$, Zongyu Li$^{1}$, and Homa Alemzadeh$^{1}$% <-this % stops a space
% \thanks{This work was partially supported by the Engineering-in-Medicine center at the University of Virginia and by the National Science Foundation under Grant No. 1842490 and 2146295.}% <-this % stops a space
\thanks{$^{1}$Kay Hutchinson, Zongyu Li, and Homa Alemzadeh are with the Department of Electrical and Computer Engineering, University of Virginia, Charlottesville, VA 22903 USA {\tt\small \{kch4fk, zl7qw, ha4d\}@virginia.edu}}%
\thanks{$^2$Ian Reyes was with the Department of Computer Science, University of Virginia, Charlottesville, VA 22903 USA. He is now with IBM. {\tt\small ir6mp@virginia.edu}}%
%\thanks{Kay Hutchinson led the development of the framework, methods, and models. Ian Reyes contributed to collecting and analyzing the dataset. Zongyu Li assisted in developing the TCN model. Homa Alemzadeh provided critical feedback, and supervised the project, analysis, and writing.}
\thanks{*Corresponding author}
\thanks{© 2023 IEEE.  Personal use of this material is permitted.  Permission from IEEE must be obtained for all other uses, in any current or future media, including reprinting/republishing this material for advertising or promotional purposes, creating new collective works, for resale or redistribution to servers or lists, or reuse of any copyrighted component of this work in other works.}
}

\maketitle

%%=============================================================%%
%% Prefix	-> \pfx{Dr}
%% GivenName	-> \fnm{Joergen W.}
%% Particle	-> \spfx{van der} -> surname prefix
%% FamilyName	-> \sur{Ploeg}
%% Suffix	-> \sfx{IV}
%% NatureName	-> \tanm{Poet Laureate} -> Title after name
%% Degrees	-> \dgr{MSc, PhD}
%% \author*[1,2]{\pfx{Dr} \fnm{Joergen W.} \spfx{van der} \sur{Ploeg} \sfx{IV} \tanm{Poet Laureate} 
%%                 \dgr{MSc, PhD}}\email{iauthor@gmail.com}
%%=============================================================%%

% \author*[1]{\fnm{Kay} \sur{Hutchinson}}\email{kch4fk@virginia.edu}

% \author[2,3]{\fnm{Ian} \sur{Reyes}}\email{ir6mp@virginia.edu}
% %\equalcont{These authors contributed equally to this work.}

% \author[1]{\fnm{Zongyu} \sur{Li}}\email{zl7qw@virginia.edu}

% \author[1,2]{\fnm{Homa} \sur{Alemzadeh}}\email{ha4d@virginia.edu}

% \affil*[1]{\orgdiv{Department of Electrical and Computer Engineering}, \orgname{University of Virginia}, \orgaddress{\city{Charlottesville}, \postcode{22903}, \state{VA}, \country{USA}}}

% \affil[2]{\orgdiv{Department of Computer Science}, \orgname{University of Virginia}, \orgaddress{\city{Charlottesville}, \postcode{22903}, \state{VA}, \country{USA}}}

% \affil[3]{\orgname{IBM}, \orgaddress{\city{RTP}, \postcode{27709}, \state{NC}, \country{USA}}}

%%==================================%%
%% sample for unstructured abstract %%
%%==================================%%

\begin{abstract}
% Surgical activity recognition is critical for skill assessment, autonomy, and error detection. 
\textcolor{black}{Fine-grained activity recognition enables explainable analysis of procedures for skill assessment, autonomy, and error detection in robot-assisted surgery.}
However, existing recognition models suffer from the limited availability of annotated datasets with both kinematic and video data %\textcolor{black}{from the robot}
and an inability to generalize to unseen subjects and tasks. %\textcolor{black}{Models that leverage multiple sources of data show promise in improving activity recognition. But, kinematic data is critical for autonomy and error detection since it does not suffer from common camera issues such as occlusions and lens contamination.} 
\textcolor{black}{Kinematic data from the surgical robot is particularly critical for safety monitoring and autonomy, as it is unaffected by common camera issues such as occlusions and lens contamination.}
We leverage an aggregated dataset of six dry-lab surgical tasks \textcolor{black}{from a total of 28 subjects} to train activity recognition models at the gesture and motion primitive \textcolor{black}{(MP)} levels and for separate robotic arms using only kinematic data.  %accuracy, edit score, and mean average precision (mAP). 
The models are evaluated %in comparison to state-of-the-art gesture and \textcolor{black}{action} recognition models 
using the LOUO \textcolor{black}{(Leave-One-User-Out)} %method and 
\textcolor{black}{and our proposed LOTO (Leave-One-Task-Out) cross validation methods to assess their ability to generalize to unseen users and tasks respectively.} % with our proposed LOTO (Leave-One-Task-Out) cross validation method.} %our proposed LOTO (Leave-One-Task-Out) cross validation method to assess a model's ability to generalize to an unseen task.  
%Our models achieve same or slightly better performance (average accuracy and mean average precision (mAP)) than the state-of-the-art in recognizing gestures (from JIGSAWS dataset) and similar motion primitives (Grasp and Pull/Retract).
% which can guide the development of models with respect to the type of data needed to train them
%Gesture recognition models outperform \textcolor{black}{MP} recognition models \textcolor{black}{in terms of accuracy and edit score}.
\textcolor{black}{Gesture recognition models achieve \textcolor{black}{higher} accuracies and edit scores than MP recognition models.} %But motion primitives enable the aggregation of data, generalization across different tasks and datasets, and separate left and right analysis. %the generation of separate left and right transcripts.
%Segmentation of tasks to motion primitives enables the generation of separate left and right transcripts and significantly improves LOTO performance. 
But, %the aggregation of data from different tasks and datasets 
using \textcolor{black}{MP}s enables the training of models that can generalize better to unseen tasks. % than gesture recognition models. 
Also, higher \textcolor{black}{MP} recognition accuracy can be achieved by training separate models for the left and right robot arms.
For task-generalization, \textcolor{black}{MP} recognition models perform best if trained on \textcolor{black}{similar} tasks and/or tasks from the same dataset. % Models for tasks with different context perform best when data from all other tasks is used for their training.

%Modeling surgical tasks with motion primitives enables the aggregation of different datasets for training recognition models that can generalize better to unseen tasks than models trained at the gesture level. 
%Our formal framework and aggregate dataset can support the development of models and algorithms for surgical process analysis, skill assessment, error detection, and autonomy.

\end{abstract}


%%\pacs[JEL Classification]{D8, H51}

%%\pacs[MSC Classification]{35A01, 65L10, 65L12, 65L20, 65L70}

%\maketitle

\section{Introduction}
\label{sec:introduction}
\textcolor{black}{In robot-assisted surgery (RAS)}, modeling and analysis at \textcolor{black}{the gesture and action} levels of the surgical hierarchy \cite{neumuth2011modeling, lalys2014surgical} is performed to gain a \textcolor{black}{better} understanding of surgical activity and improve skill assessment \cite{tao2012sparse, varadarajan2009data}, error detection \cite{yasar2019context, yasar2020real,hutchinson2022analysis,li2022runtime}, and autonomy \cite{ginesi2021overcoming}. Towards these applications, automated segmentation and classification of surgical workflow has been an active area of research \cite{valderrama2022towards}. %\textcolor{black}{%there has been much work at the gesture and action levels as summarized in 
\cite{van2021gesture} and \cite{hutchinson2023compass} \textcolor{black}{provide comprehensive} summaries of the recent works  at the gesture and action levels. %While gesture recognition has been done with kinematic and/or video data \cite{van2021gesture}, recent work on action triplet recognition has mainly focused on video data of real surgical procedures \cite{nwoye2022rendezvous, li2022sirnet}. 
% It is shown that models trained for multi-granularity recognition perform better than models for activity recognition alone \cite{huaulme2021micro}, so incorporating knowledge of higher surgical process levels can improve finer-grained recognition. 
However, previous works and comparisons \textcolor{black}{among} them have been restricted by differing gesture definitions \cite{van2021gesture} and limited diversity in the numbers of subjects, trials, and \textit{tasks} \textcolor{black}{across the existing datasets}. 

% Grammar graphs \cite{ahmidi2017dataset} model the decomposition of tasks into gestures and \cite{hutchinson2023compass} models tasks with context and motion primitives, but there are not yet models relating fine-grained actions to existing gestures. Such relationships would support surgical process modeling, comparative analysis across datasets and models, and analyses to understand and detect errors in robotic surgery \cite{hutchinson2022analysis, li2022runtime}.

% While gesture recognition has been done with kinematic and/or video data \cite{van2021gesture}, recent work on action triplet recognition has mainly focused on video data of real surgical procedures \cite{nwoye2022rendezvous, li2022sirnet}. 
% But, kinematic data is important for improved recognition accuracy using multi-modal data~\cite{qin2020temporal, qin2020davincinet} and error detection \cite{van2021gesture, yasar2020real, li2022runtime, hutchinson2022analysis} since it is unaffected by common camera issues such as noise, occlusions, lens contamination, and smoke~\cite{yasar2019context, yong2016impact, allers2016evaluation}. 

Recent works in gesture recognition have each defined their own sets of gestures for their own datasets \cite{dipietro2019segmenting, goldbraikh2022using, menegozzo2019surgical, gonzalez2021dexterous, de2021first} with limited overlap between gestures. % and% as well as proposing and recognizing 
On the other hand, works on recognition of fine-grained surgical actions focus on action triplets \textcolor{black}{(verb, instrument, tissue/object)}~\cite{meli2021unsupervised, li2022sirnet, nwoye2022rendezvous}, representing  surgical instrument and tissue interactions in endoscopic videos. %But the combinations of verbs, instruments, and targets for action triplets can grow exponentially compared to a more limited number of gestures based on descriptive definitions. %Also, these works consider verb recognition separately from recognition of the full action triplet and use only video data. In comparison, we focus on verb recognition using only kinematic data since the tools and objects would need to be inferred from image data. 
\textcolor{black}{W}hile gesture recognition has been done with kinematic and/or video data \cite{van2021gesture}, recent work on action triplet recognition has mainly focused on video data of surgical procedures \cite{nwoye2022rendezvous, li2022sirnet}. \textcolor{black}{To leverage finer-grained action recognition in safety monitoring and autonomy applications, \textcolor{black}{in this paper} we examine verb-only predictions based on kinematic data.} 
% But, kinematic data is important for improved recognition accuracy using multi-modal data~\cite{qin2020temporal, van2022gesture} and for error detection \cite{van2021gesture, yasar2020real, li2022runtime, hutchinson2022analysis}, since it is unaffected by common camera issues such as occlusions, lens contamination, and smoke~\cite{yasar2019context, yong2016impact, allers2016evaluation}. For this reason, we use kinematic data in this work as a step towards a context-aware safety monitoring system.
\textcolor{black}{Kinematic data is particularly important for safety analysis~\cite{li2022runtime, hutchinson2022analysis}, error detection~\cite{van2021gesture, yasar2020real}, and improved recognition accuracy using multi-modal data~\cite{qin2020temporal, van2022gesture}, since it is unaffected by common camera issues such as occlusions, lens contamination, and smoke~\cite{yasar2019context, yong2016impact, allers2016evaluation}. Plus, using fewer data types can reduce computational cost and enable real-time applications \cite{shi2022recognition}.} %For this reason, we use kinematic data in this work as a step towards a context-aware safety monitoring system.}

To address the challenge of limited datasets, Hutchinson et al. presented a new dataset, called COMPASS \cite{hutchinson2023compass}, which aggregates six dry-lab surgical training tasks from the JIGSAWS \cite{gao2014jhu}, DESK \cite{gonzalez2021dexterous}, and ROSMA \cite{rivas2021surgical} datasets by providing standardized \textit{context} and \textit{motion primitive (MP)} labels for all the tasks. \textcolor{black}{MPs are a standardized set of actions (e.g., push) whose execution results in changes of surgical context, which is comprised of important state variables describing physical status and interactions of tools and tissues/objects (e.g., needle in tissue).} % that comprise the surgical context. %This dataset includes both video and kinematic data and separate transcripts for the left and right robot arms. %JIGSAWS and DESK also have gesture labels from their original datasets. 
% Also, some of the tasks in the dataset share similar semantics and physical context and have gesture labels from their original datasets.
\textcolor{black}{Some of the tasks in the dataset share similar objects and goals enabling their aggregation and comparison.} %semantics and physical context and have gesture labels from their original datasets.
%\cite{hutchinson2022analysis} also labeled the Suturing and Needle Passing tasks of the JIGSAWS dataset for executional errors and found that the definitions of some gestures differed between tasks and that parts of the discrepancy were important for differentiating between erroneous and non-erroneous performances of that gesture. 
The standardized labels in COMPASS can support aggregated analysis of datasets and combining data from contextually similar tasks for improved activity recognition and error detection~\cite{van2021gesture, hutchinson2022analysis, li2022runtime}. 

\textcolor{black}{In this paper, we use the COMPASS dataset to study the effect of label granularity on activity recognition performance and generalization across users, tasks, and datasets for RAS with a case study of Temporal Convolutional Networks (TCN)  \cite{lea2016temporal}.} Specifically, we make the following contributions:

%Another recent work modeled surgical processes using statecharts with surgemes and triggers \cite{falezza2021modeling}. %Related work on gestures and actions with label definitions are presented in Section \ref{sect:related_work} and Tables \ref{tab:related_work_gestures} and \ref{tab:related_work_actions}. In all these works, the granularity of the actions %between them 
%can vary from the sub-gesture to the task level (see  Figure~\ref{fig:hierarchy}). %A recent approach to surgical process modeling used statecharts with four surgemes summarized as idle, move, rotate, and open/close grasper \cite{falezza2021modeling}. 
%Yet, there is still no formal framework that defines a standard set of surgical actions and their relations to gestures and tasks, which would enable direct comparisons between these works, their datasets, and their recognition models. 

%Our contributions are as follows:

\begin{itemize}
\item We compare the performance of \textcolor{black}{existing} activity recognition \textcolor{black}{models in a case study of TCN} using only kinematic data at different levels of the surgical hierarchy, specifically, the gesture and motion primitive levels, and for separate left and right sides of the robot vs. both sides combined. %Separate  motion primitive recognition models for the left and right sides of the robot generally perform better than a single model for both sides combined.

\item We introduce the Leave-One-Task-Out (LOTO) cross validation method to measure the ability of surgical activity recognition models to generalize to an unseen task, since current datasets do not include all of the surgical tasks that a model may see when it is deployed. 

\item We perform the first evaluation of a surgical activity recognition model trained on multiple tasks with data combined from different datasets by comparing model performance using the existing LOUO method as well as our proposed LOTO cross validation method. 

\end{itemize}

The insights from our analysis can guide the development of future surgical activity recognition and error detection models. The aggregated dataset and code to train and evaluate the recognition models are publicly available at \url{https://github.com/UVA-DSA/COMPASS}.% to facilitate further research and collaboration in this area. 

\section{Background}

\begin{figure}[b!]
    \vspace{-2em}
    \centering
    % trim left, bottom, right, top
    \includegraphics[trim = 0in 4.3in 7in 0in, width=0.6\textwidth]{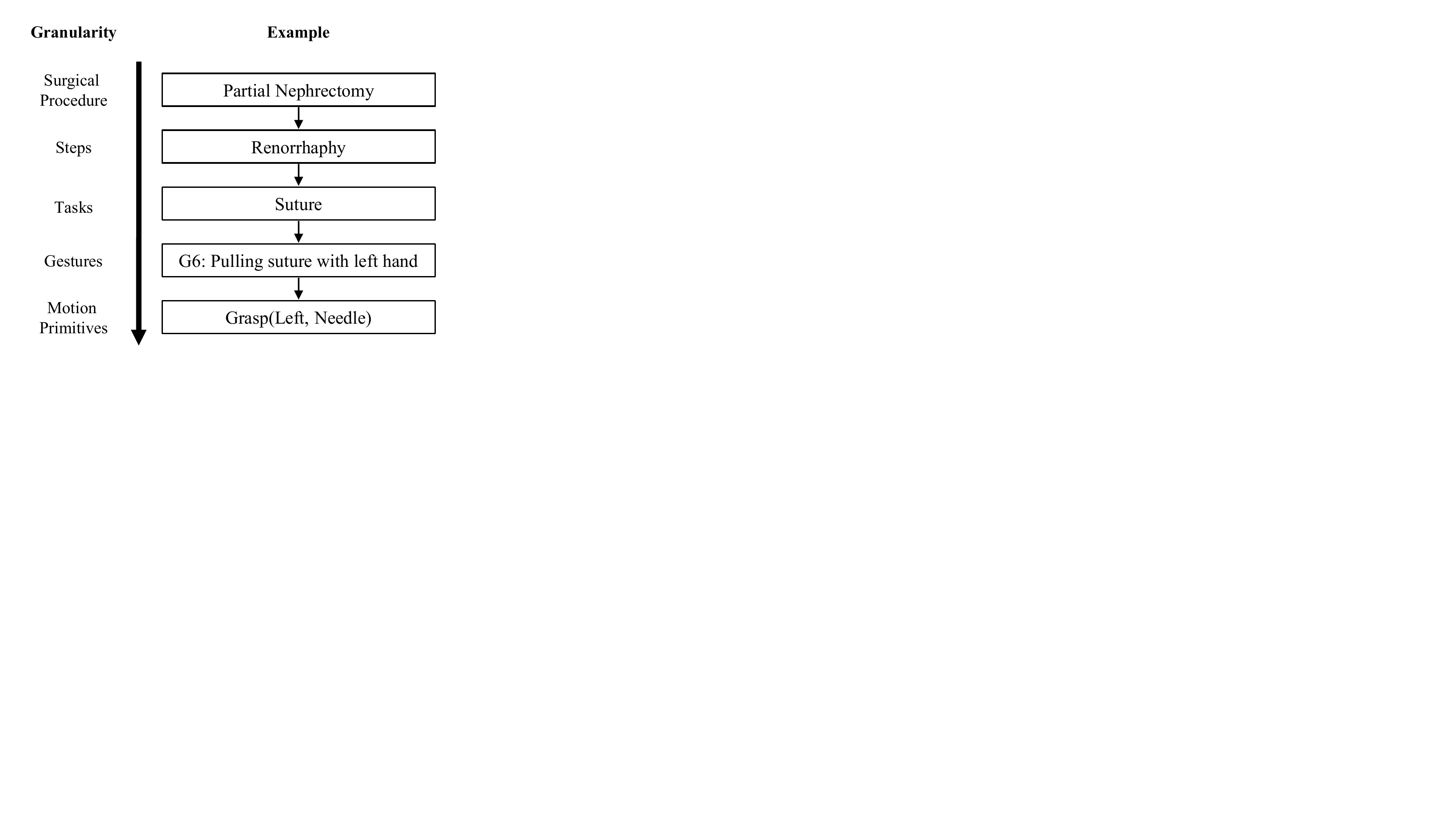}
    \caption{Surgical Hierarchy. Adapted from \cite{hutchinson2022analysis}}
    \label{fig:hierarchy}
\end{figure}

\subsection{Levels of Granularity in Surgical Procedures}
Surgical process modeling \cite{neumuth2011modeling, lalys2014surgical} decomposes surgical procedures into smaller units such as steps, tasks, gestures, and actions as shown in Figure \ref{fig:hierarchy}. \textcolor{black}{We refer to units at any level of the surgical hierarchy as ``activities".}
Gestures are defined as ``intentional surgical activit[ies] resulting in a perceivable and meaningful outcome" %the smallest surgical motion gesture that encapsulates a specific intent", 
(e.g., pushing needle through tissue) \cite{gao2014jhu} and usually include the semantics of both the activity and the underlying physical context in their definition. 
\textcolor{black}{We also consider surgical actions (i.e., the verbs of action triplets~\cite{nwoye2022rendezvous, neumuth2006acquisition}) which are atomic units of activity or lower level motions (e.g., grasp, push) based on kinematic data, but without the semantics of physical context or the types and status of interacting tools and objects/tissues (e.g., needle through tissue) based on video data \cite{neumuth2011modeling}.}
%On the other hand, surgical actions (i.e., the verbs of action triplets~\cite{nwoye2022rendezvous, neumuth2006acquisition}) are atomic units of activity or lower level motions (e.g., grasp, push) without capturing the semantics of physical context (e.g., needle through tissue) \cite{neumuth2011modeling}. 
%Much work has been done in developing models for gesture recognition mostly using data from the Suturing task in the JIGSAWS dataset as summarized in \cite{van2021gesture}. Surgical gesture recognition has applications in skill assessment \cite{forestier2012classification, wang2021towards}, error detection \cite{yasar2020real, hutchinson2022analysis, li2022runtime}, and autonomy \cite{de2021first}. 

\begin{figure}[t!]
    \centering
    \begin{minipage}{0.49\textwidth}
        \centering
        % trim left, bottom, right, top
        \includegraphics[trim = 0in 5in 8in 0in, clip, 
    width=\textwidth]{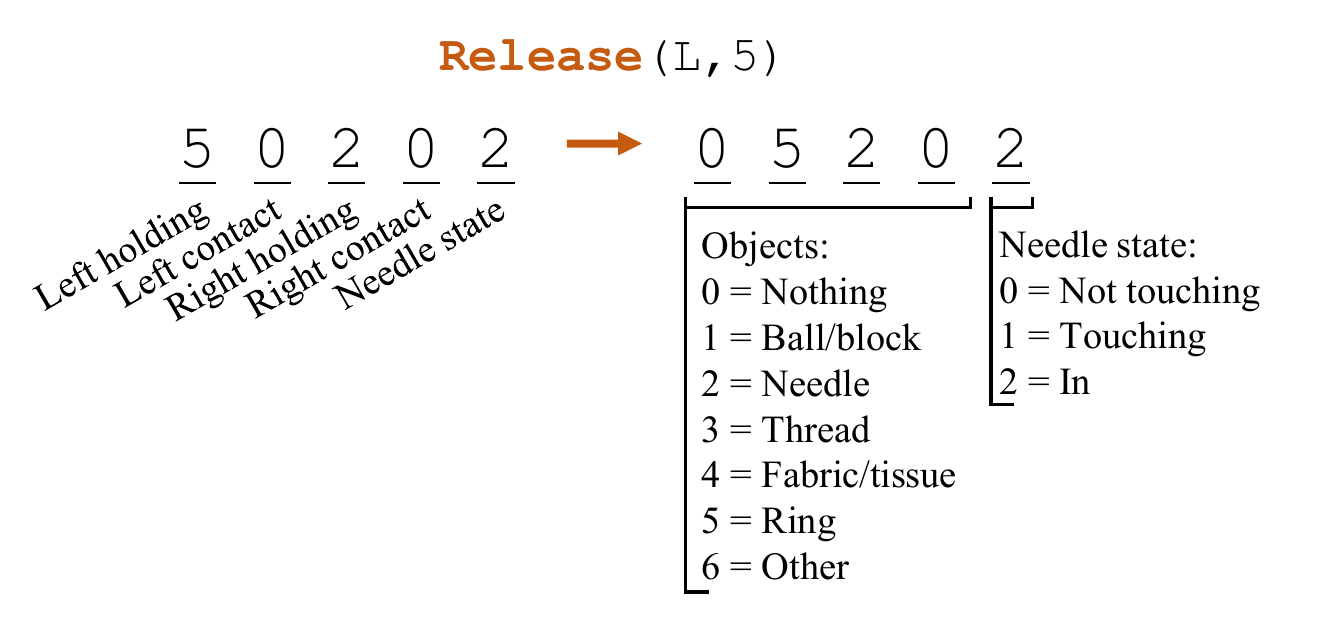}
        \vfill
    \end{minipage}
    \hfill 
    \vspace{-0.75em}
    \caption{Context states and object encodings for a ``Release" motion primitive from the Needle Passing task \cite{hutchinson2023compass}.} 
    \label{fig:context}
    \vspace{-0.5em}
\end{figure}

%Previous works \cite{yasar2020real} and \cite{van2021gesture} have provided baseline gesture recognition model accuracies for various methods including LSTM \cite{yasar2020real}, TCN \cite{lea2016temporal}, SDSDL \cite{sefati2015learning}, and SC-CRF \cite{lea2015improved} that were trained and evaluated on the JIGSAWS dataset using only patient-side kinematic data. The SDSDL and SC-CRF models provide baseline accuracies for state-of-the-art models as shown in Table \ref{tab:LOUO}. 
Existing activity recognition models have been mostly task-specific and restricted to specific datasets and gesture definitions. For example, the majority of previous works have used the JIGSAWS dataset and gesture definitions~\cite{gao2014jhu}. 
To address this, \cite{hutchinson2023compass} defined a finer-grained set of motion primitives (MPs) as generalizable surgical actions to enable comparative analysis between tasks and datasets. MPs are similar in granularity and definition to the action triplets defined by \cite{nwoye2022rendezvous}. % and others, but were mainly defined for the dry-lab surgical training tasks. %Each MP is defined as a modular and programmable unit that causes changes in the context representing interactions between tool and objects in the scene \cite{hutchinson2023compass}. 
%Table \ref{tab:MPs} shows the set of MPs and the corresponding verbs with similar action semantics in the action triplet definitions by  \cite{nwoye2022rendezvous}.  
Each MP consists of a verb (e.g., Grasp), the tool that is used (e.g., left grasper), and the object with which the tool interacts (e.g., needle). The left and right graspers are abbreviated as `L' and `R', and the object encodings are shown in Figure \ref{fig:context} for an example \textcolor{black}{MP} and physical context.
\textcolor{black}{Table \ref{tab:LOUO_mAPs_all} shows the set of MPs and the number of samples in each MP class and task.}
% Table \ref{tab:MPs} shows the set of MPs and the corresponding verbs with similar action semantics in the action triplet definitions~\cite{nwoye2022rendezvous, valderrama2022towards, li2022sirnet}.
% but the COMPASS dataset also contains kinematic data to support the development of autonomous robots and error detection models.

\begin{figure}[t!]
    % \vspace{-1em}
    \centering
    % \begin{minipage}[b]{.475\linewidth}
    \begin{subfigure}{0.14\textwidth}
        \centering
        % trim left, bottom, right, top
        \includegraphics[trim = 3in 0in 3in 0.5in, clip, width=\textwidth]{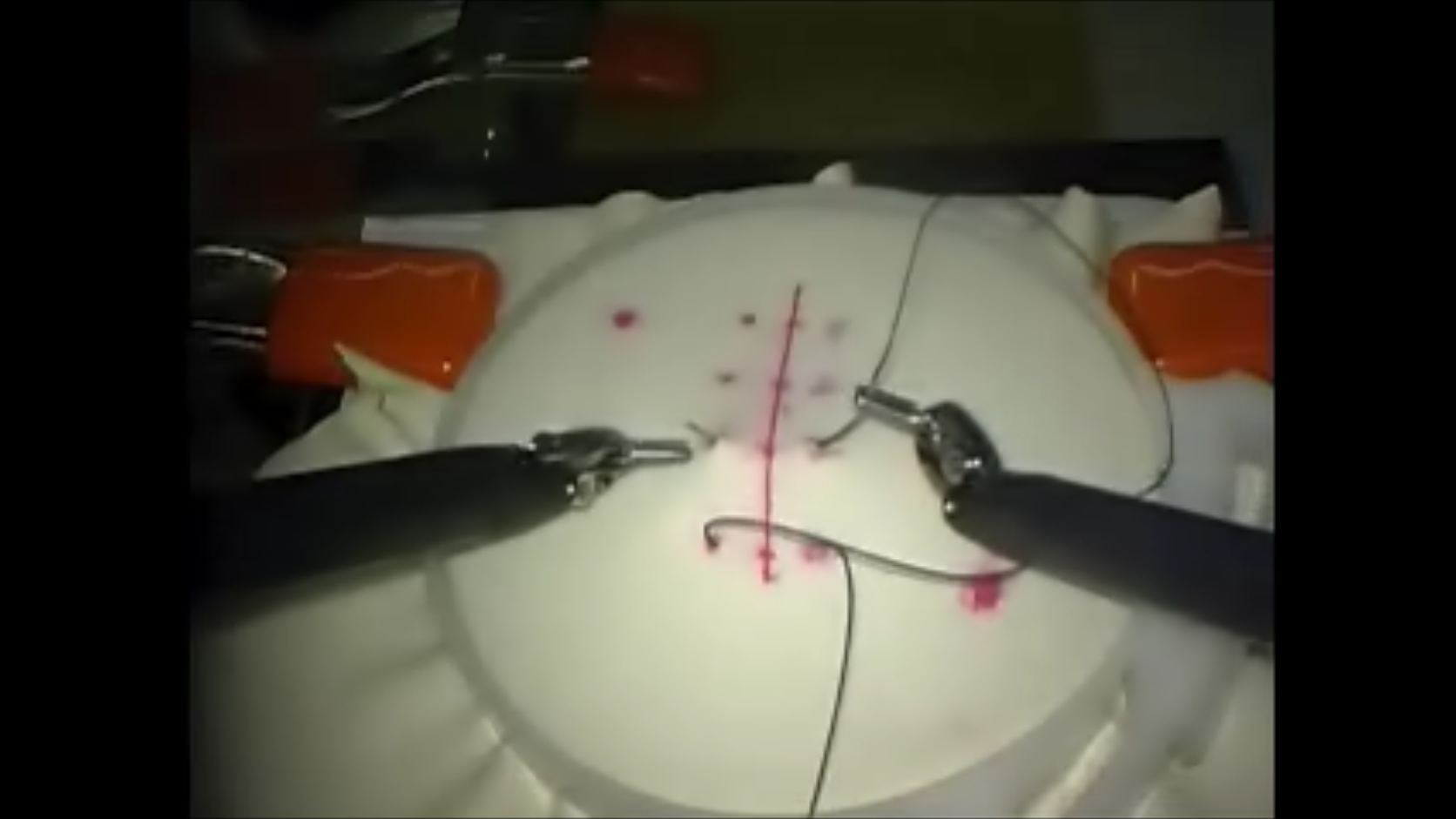}
        \caption{Suturing \\ \centering (S)}
        \label{fig:S_frame}
    \end{subfigure}
    \begin{subfigure}{0.14\textwidth}
        \centering
        % trim left, bottom, right, top
        \includegraphics[trim = 3in 0.25in 3in 0.25in, clip, width=\textwidth]{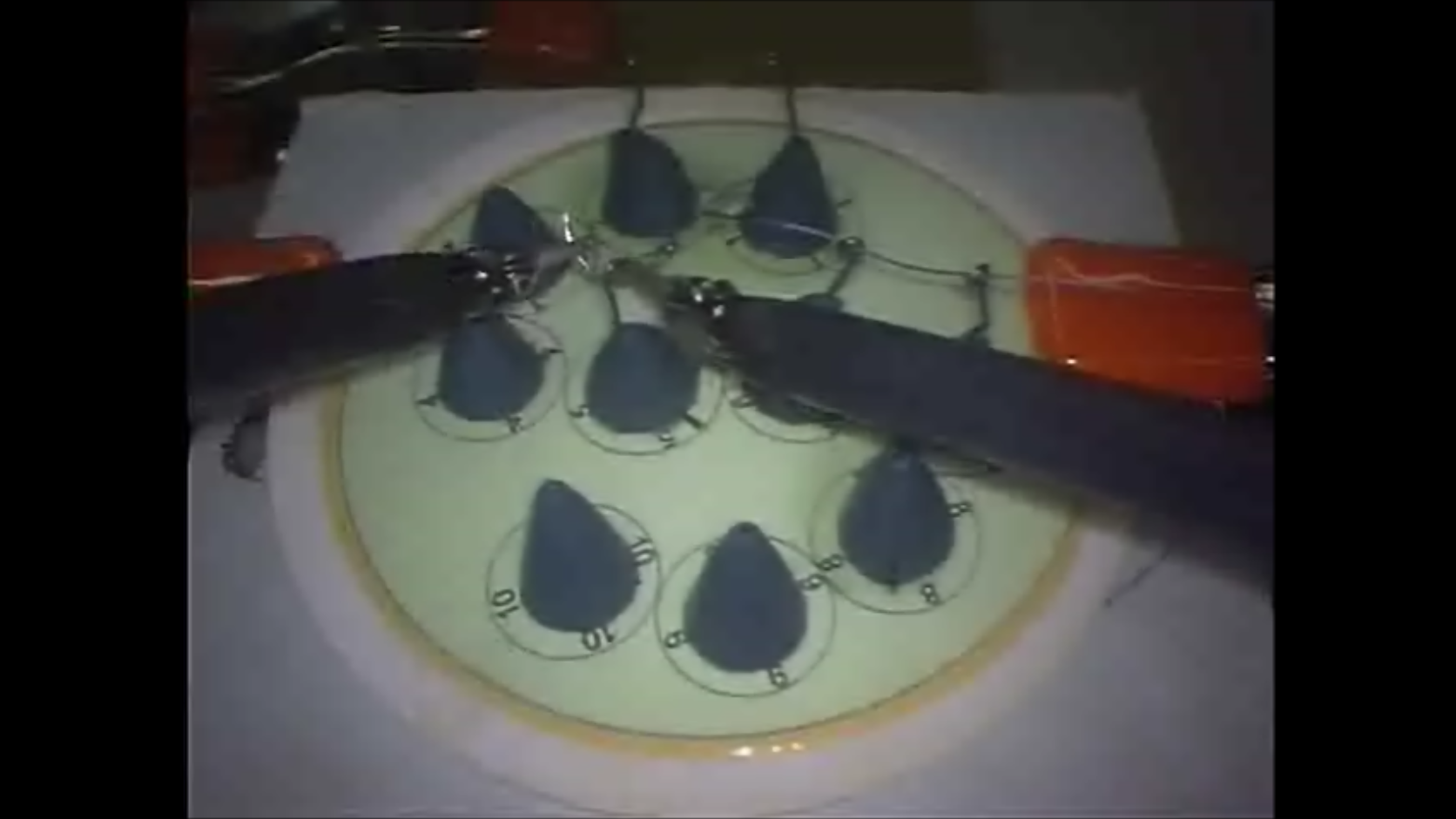}
        \caption{Needle Passing \\ \centering (NP)}
        \label{fig:NP_frame}
    \end{subfigure}
    \begin{subfigure}{0.14\textwidth}
        \centering
        % trim left, bottom, right, top
        \includegraphics[trim = 3in 0.25in 3in 0.25in, clip, width=\textwidth]{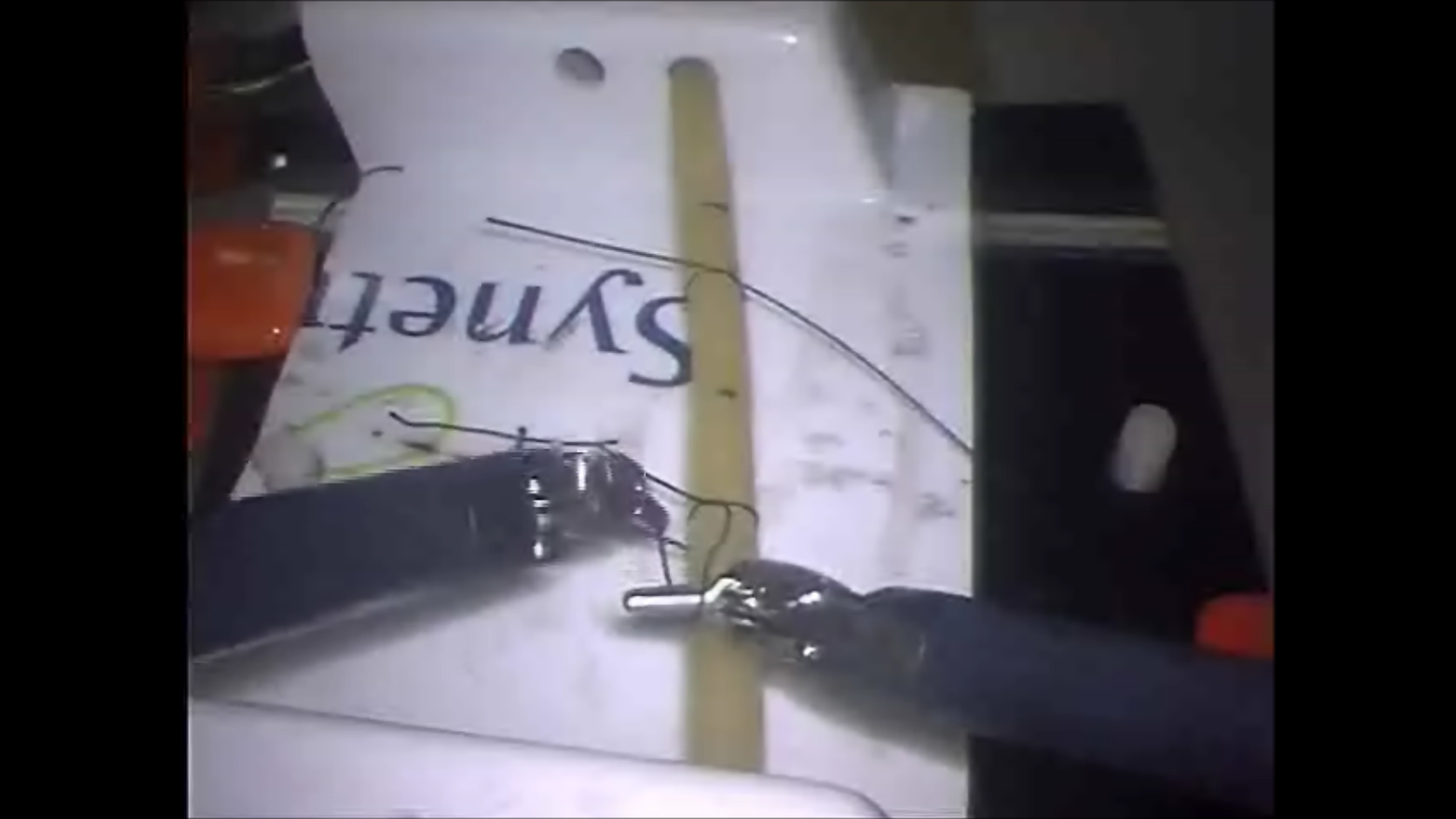}
        \caption{Knot Tying \\ \centering (KT)}
        \label{fig:KT_frame}
    \end{subfigure}
    % \end{minipage}
    % \begin{minipage}[b]{.475\linewidth}
    \begin{subfigure}{0.14\textwidth}
        \centering
        % trim left, bottom, right, top
        \includegraphics[trim = 3in 0in 3in 0.5in, clip, width=\textwidth]{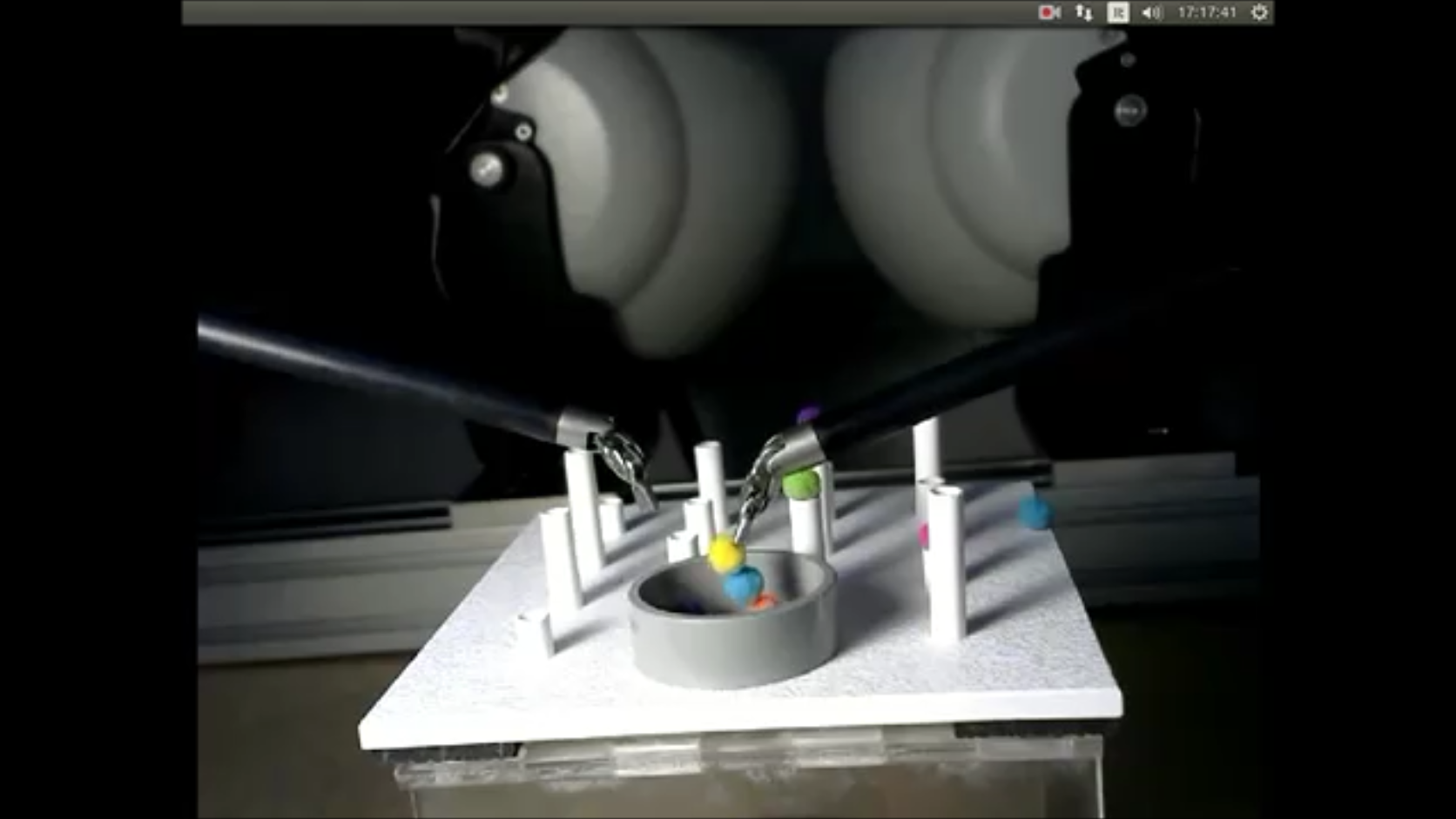}
        \caption{Pea on a Peg \\ \centering (PoaP)}
        \label{fig:PoaP_frame}
    \end{subfigure}
    \begin{subfigure}{0.14\textwidth}
        \centering
        % trim left, bottom, right, top
        \includegraphics[trim = 3in 0in 3in 0.5in, clip, width=\textwidth]{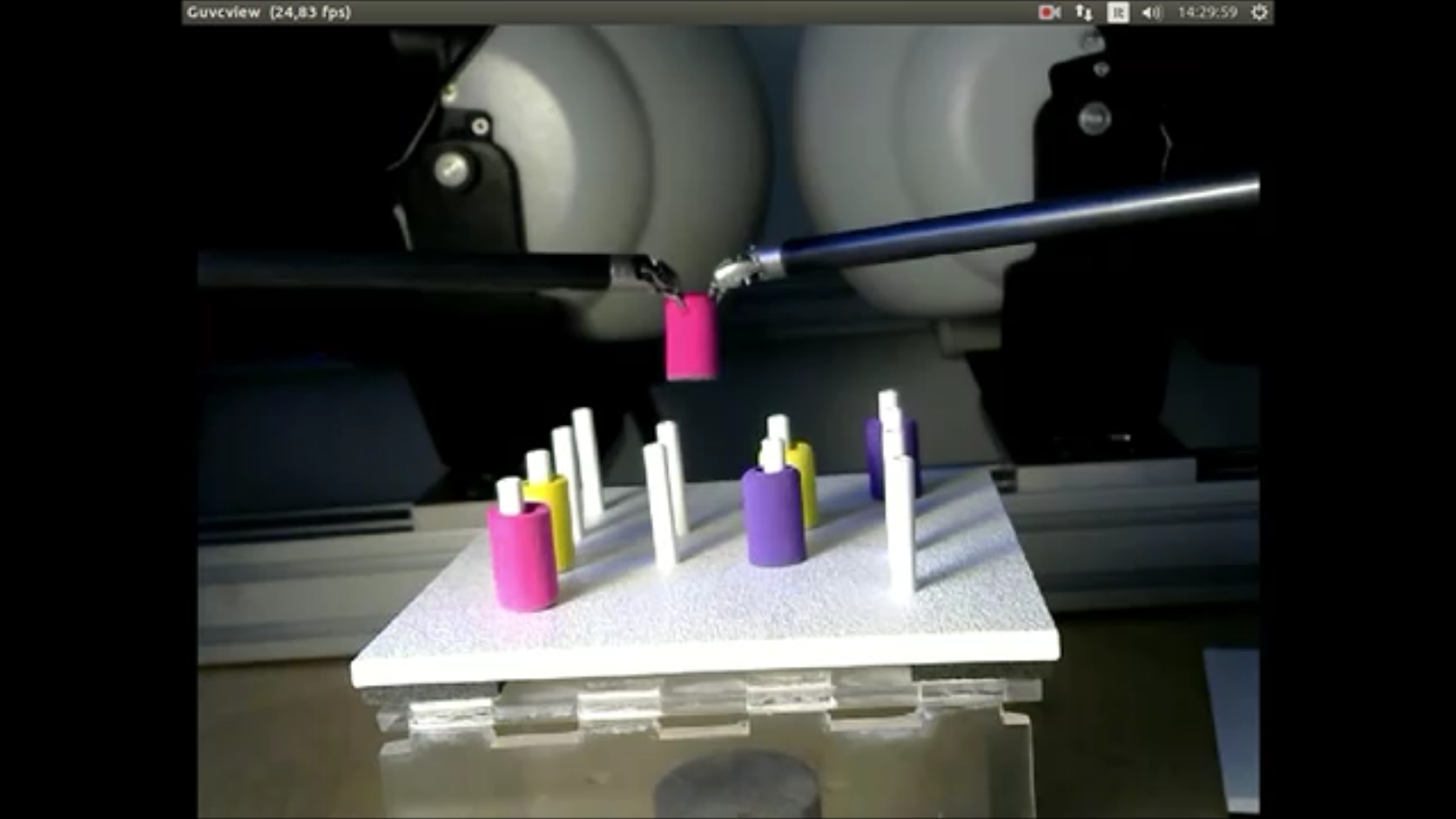}
        \caption{Post and Sleeve \\ \centering (PaS)}
        \label{fig:PaS_frame}
    \end{subfigure}
    \begin{subfigure}{0.14\textwidth}
        \centering
        % trim left, bottom, right, top
        \includegraphics[trim = 3in 0.5in 3in 0in, clip, width=\textwidth]{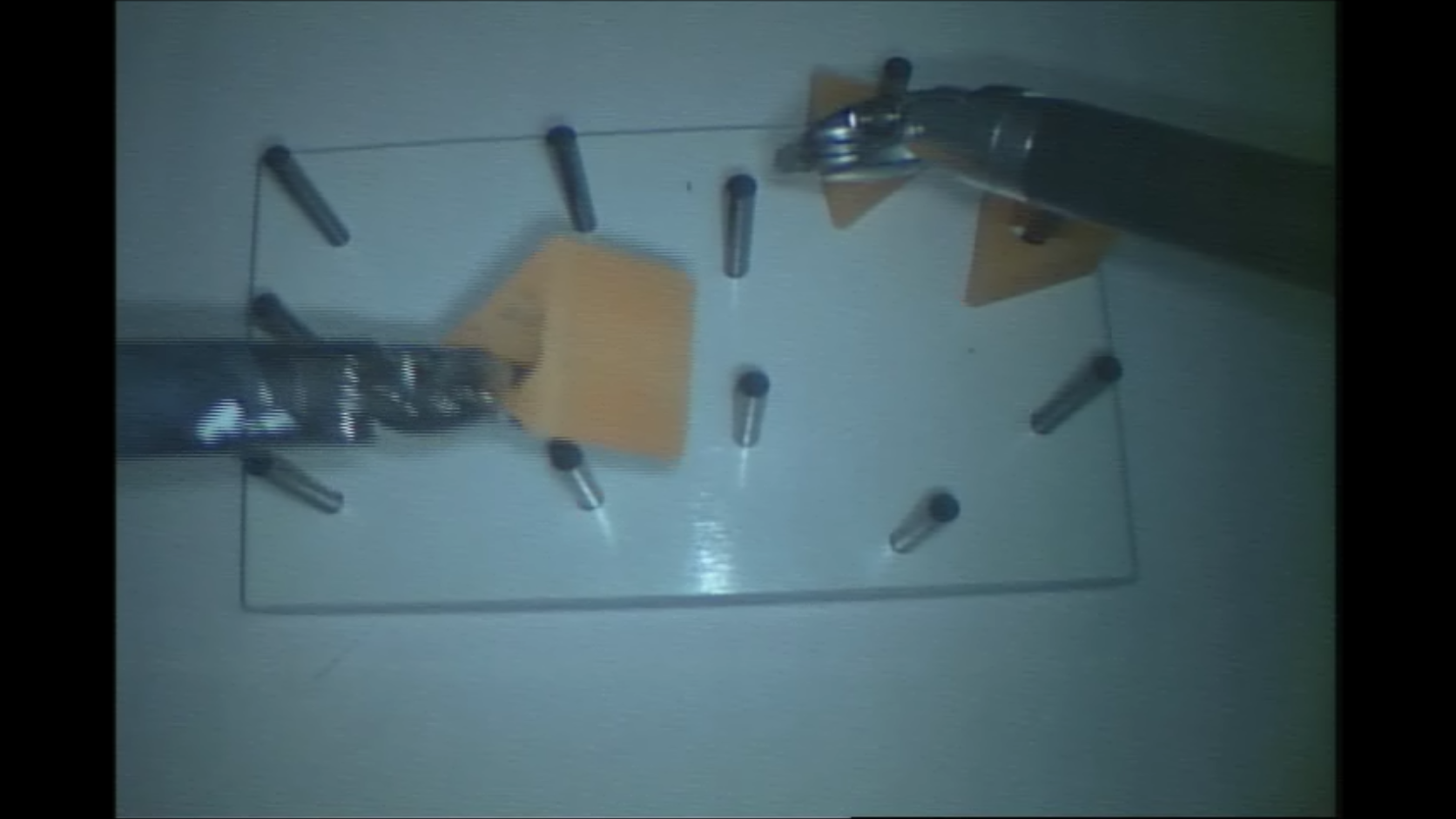}
        \caption{Peg Transfer \\ \centering (PT)}
        \label{fig:PT_frame}
    \end{subfigure}
    % \end{minipage}
    \vspace{-0.5em}
    \caption{COMPASS \textcolor{black}{tasks: S, NP, and KT from JIGSAWS \cite{gao2014jhu}; PoaP and PaS from ROSMA \cite{rivas2021surgical}; PT from DESK \cite{gonzalez2021dexterous}.}}
    \label{fig:tasks}
    \vspace{-1.75em}
\end{figure}

\subsection{COMPASS Dataset}
\begin{figure}[t!]
    \centering
    % \begin{subfigure}{0.47\textwidth}
    %     \centering
    %     % trim left, bottom, right, top
    %     \includegraphics[trim = 0in 2.5in 9in 0.25in, clip, 
    %     width=\textwidth]{Figures/NP_graph.pdf}
    %     \caption{}
    %     \label{fig:NP_graph}
    % \end{subfigure}
    \begin{minipage}{0.49\textwidth}
        % \begin{subfigure}{\textwidth}
        %     \centering
        %     % trim left, bottom, right, top
        %     \includegraphics[trim = 4.4in 5in 3.5in 0.25in, clip, 
        %     width=\textwidth]{Figures/NP_graph.pdf}
        %     \caption{}
        %     \label{fig:context}
        % \end{subfigure} 
        % \vfill
        % \begin{subfigure}{\textwidth}
            \centering
            % trim left, bottom, right, top
            %\includegraphics[trim = 4.4in 2.5in 3.75in 3in, clip, width=\textwidth]{Figures/NP_graph.pdf}
            \includegraphics[trim = 0.2in 4.7in 7.75in 0in, clip, 
            width=\textwidth]{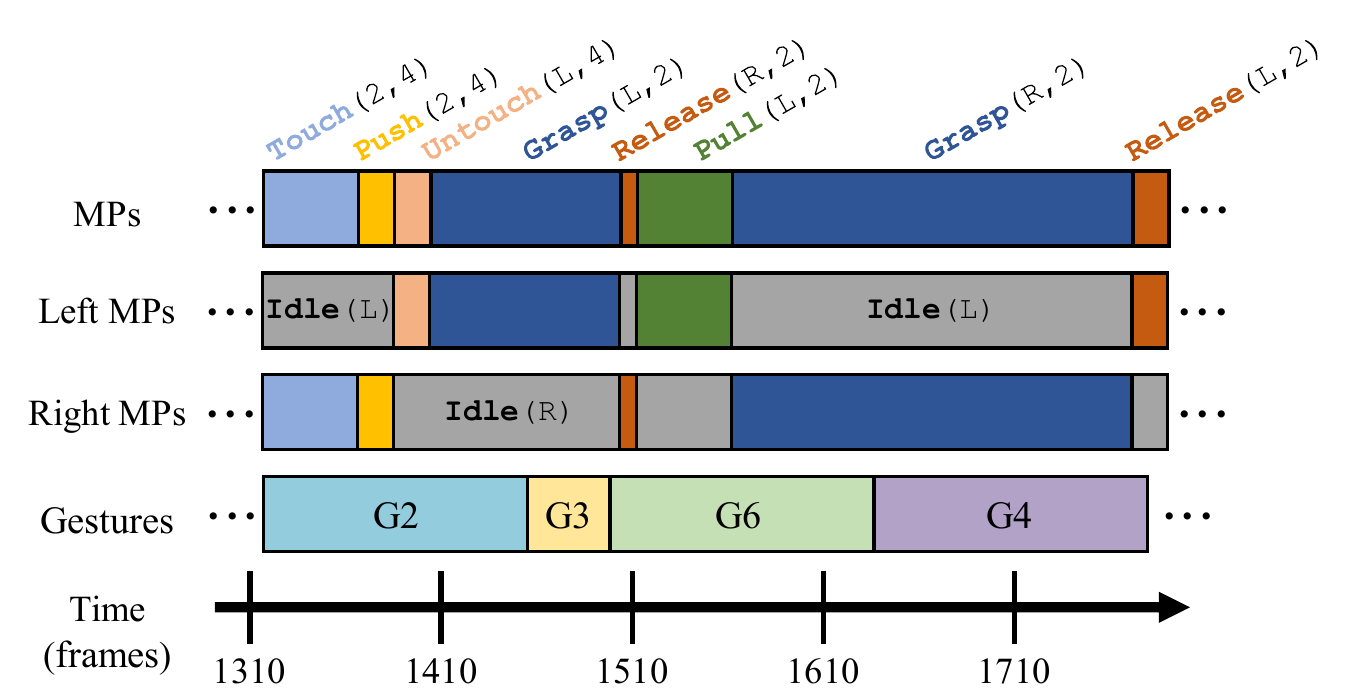}
            % \caption{}
            % \label{fig:MPsandgestures}
        % \end{subfigure}
    \end{minipage}
    \hfill 
    
    \caption{Example of alignment between MPs and gestures in a Needle Passing trial that also shows the G3 boundary discrepancy noted by \cite{hutchinson2022analysis} where the 'Push' MP is not part of G3. From \cite{gao2014jhu}, G2: positioning needle, G3: pushing needle through tissue, G4: transferring needle from left to right, G6: pulling suture with left hand. Figure best viewed in color.} %compared to gesture grammar graph from \cite{ahmidi2017dataset}}
    \label{fig:MPsandgestures}
    \vspace{-2em}
\end{figure}

We use the COMPASS dataset \cite{hutchinson2023compass} \textcolor{black}{since it has different dry-lab tasks from multiple datasets and kinematic data from da Vinci surgical robots with which to train our surgical activity recognition models. We compare the performance of these models at the gesture and MP granularities.}
%The COMPASS dataset contains kinematic and video data for 39 trials of Suturing (S), 28 trials of Needle Passing (NP), and 36 trials of Knot Tying (KT) performed by eight subjects from the JIGSAWS dataset \cite{gao2014jhu}; 47 trials of Peg Transfer (PT) performed by eight subjects from the DESK dataset \cite{gonzalez2021dexterous}; and 65 trials of Post and Sleeve (PaS), and 71 trials of Pea on a Peg (PoaP) performed by 12 subjects from the ROSMA dataset \cite{rivas2021surgical} (see Figure \ref{fig:tasks}).
\textcolor{black}{The COMPASS dataset contains kinematic and video data \textcolor{black}{at 30 Hz} for a total of six tasks from three different datasets as described in Table \ref{tab:COMPASS_dataset}. The tasks are: Suturing (S), Needle Passing (NP), Knot Tying (KT), Peg Transfer (PT), Post and Sleeve (PaS), and Pea on a Peg (PoaP) as shown in Figure \ref{fig:tasks}.}
Context and MP labels are present for all trials, but gesture labels are only available for trials in the JIGSAWS and DESK datasets. %Separate MP transcripts for the left and right sides of the robot were also included in the COMPASS dataset. 
\textcolor{black}{To generate separate left and right label sets, MPs performed by each arm of the robot are split into new transcripts. Also, an 'Idle' MP is defined and used to fill the gaps created by the separation so that every kinematic sample has a label.}
An example segment of a Needle Passing trial with each label type is shown in Figure \ref{fig:MPsandgestures}. This also shows the discrepancy in the G3 boundary noted by \cite{hutchinson2022analysis} where the Push(Needle, Ring) MP is in G2 rather than G3.
\begin{comment}
\begin{table}[]
    \centering
    \caption{\textcolor{black}{Number of subjects and trials for each task in the COMPASS dataset: Suturing (S), Needle Passing (NP), Knot Tying (KT), Peg Transfer (PT), Post and Sleeve (PaS), and Pea on a Peg (PoaP).}}
    \label{tab:COMPASS_dataset}
    \begin{tabular}{ccccccc}
        \toprule
         & \multicolumn{3}{c}{JIGSAWS \cite{gao2014jhu}} & DESK \cite{madapana2019desk} & \multicolumn{2}{c}{ROSMA \cite{rivas2021surgical}} \\
         \cmidrule(lr){2-4} \cmidrule(lr){5-5} \cmidrule(lr){6-7}
        Task & S & NP & KT & PT & PaS & PoaP \\
        \midrule
        Trials & 39 & 28 & 36 & 47 & 65 & 71 \\
        Subjects & 8 & 8 & 8 & 8 & 12 & 12 \\
        \bottomrule
    \end{tabular}
\end{table}
\end{comment}

\begin{table}[]
    % \vspace{-1.75em}
    \vspace{0.5em}
    \centering
    \caption{\textcolor{black}{Number of subjects and trials and types of annotations for each task in the COMPASS dataset: Suturing (S), Needle Passing (NP), Knot Tying (KT), Peg Transfer (PT), Post and Sleeve (PaS), and Pea on a Peg (PoaP).}}
    \label{tab:COMPASS_dataset}
    \begin{tabular}{ccccccc}
        \toprule
         Dataset & \multicolumn{3}{c}{JIGSAWS \cite{gao2014jhu}} & DESK \cite{madapana2019desk} & \multicolumn{2}{c}{ROSMA \cite{rivas2021surgical}} \\
         \midrule 
         Tasks & S & NP & KT & PT & PaS & PoaP \\
         % \midrule
         Trials & 39 & 28 & 36 & 47 & 65 & 71 \\
         \cmidrule(lr){1-1} \cmidrule(lr){2-4} \cmidrule(lr){5-5} \cmidrule(lr){6-7}
         Subjects & \multicolumn{3}{c}{8} & 8 & \multicolumn{2}{c}{12} \\
         Gesture Labels & \multicolumn{3}{c}{\checkmark} & \checkmark & \multicolumn{2}{c}{} \\
         MP Labels & \multicolumn{3}{c}{\checkmark} & \checkmark & \multicolumn{2}{c}{\checkmark} \\
         
        \bottomrule
    \end{tabular}
    %\vspace{-1em}
\end{table}

%%%%
% The COMPASS dataset includes consensus context labels at 3 Hz and automatically generated motion primitive labels interpolated to 30 Hz for both arms of the robot so that every kinematic sample has an MP label. %Prior work has only considered one set of labels that describe the actions of both arms, however, MPs enable the generation of separate transcripts for the left and right arms. %To accomplish this, MPs performed by each arm of the robot are split into new transcripts and the 'Idle' MP is used to fill the gaps created by the separation so that every sample has a label.
% The original gesture labels from JIGSAWS and DESK datasets are synced and renamed under the new naming convention and included to promote comparisons between data and label sets.

\noindent
\begin{table*}[ht!]
\vspace{0.5em}
\begin{center}
% \begin{minipage}{\textwidth} %{174pt}
\caption{Surgical workflow segmentation models that considered multiple datasets and label granularities.} 
%\textcolor{black}{* normalized by maximum number of
%segments in any ground-truth sequence.}}
\label{tab:related_work}
\resizebox{\textwidth}{!}{
\begin{tabular}{p{2.25cm} p{1.5cm} p{1.25cm} p{1.6cm} p{1.5cm} p{3.5cm} p{4cm}}
\toprule
%Paper & Dataset & Tasks & Actions \\
Paper & Dataset & Data Type & Tasks & Label Levels & \multicolumn{2}{p{5cm}}{Best Teams/Models and Performance} \\
\midrule

\multirow{4}{2cm}{MISAW Challenge 2021 \cite{huaulme2021micro}} & \multirow{4}{2cm}{MISAW} & \multirow{4}{1.5cm}{Kin and/or Video} & \multirow{4}{2cm}{Anastomosis} & Phases & MedAIR \cite{long2021relational} & AD-Accuracy: 96.5\%  \\
\cmidrule(lr){5-7} \hfill
 & & & & Steps & MedAIR \cite{long2021relational} & AD-Accuracy: 84.0\% \\ 
 \cmidrule(lr){5-7} \hfill
 & & & & Activities & NUSControl Lab and UniandesBCV & AD-Accuracy: $ \sim$64\% \\
 \cmidrule(lr){5-7} \hfill
  & & & & Multigranularity & NUSControl Lab & AD-Accuracy: $ \sim $72\% \\
%\hline
\midrule

\multirow{2}{2.5cm}{HeiChole benchmark 2021 \cite{wagner2023comparative}} & \multirow{2}{2cm}{EndoVis 2019} & \multirow{2}{2cm}{Video} & \multirow{2}{2cm}{Laparoscopic cholecystectomy} & Phase & HIKVision and CUHK & F1 Score: $ \sim$65\% \\
\cmidrule(lr){5-7} \hfill
 & & & & Actions & Wintegral & F1 Score: 23.3\% \\
%\hline
\midrule

\multirow{3}{2cm}{Valderrama 2022 \cite{valderrama2022towards}} & \multirow{3}{2cm}{PSI-AVA} & \multirow{3}{2cm}{Video} & \multirow{3}{2cm}{Radical prostatectomy} & Phase & \multirow{3}{2cm}{TAPIR} & mAP: 56.6\% \\
\cmidrule(lr){5-5} \cmidrule(lr){7-7} \hfill
 & & & & Step & & mAP: 45.6\% \\
\cmidrule(lr){5-5} \cmidrule(lr){7-7} \hfill
 & & & & Action & & mAP: 23.6\% \\
%\hline
\midrule

\multirow{2}{2.25cm}{DiPietro 2019 \cite{dipietro2019segmenting}} & JIGSAWS & \multirow{2}{2cm}{Kinematics} & Suturing & Gestures & \multirow{2}{2cm}{GRU} & Error rate: 15.2\% Edit distance: 8.4 \\
\cmidrule(lr){2-2} \cmidrule(lr){4-5} \cmidrule(lr){7-7} \hfill
 & MISTIC-SL & & Knot Tying & Maneuvers &  & Error rate: 8.6\% Edit distance: 9.3 \\
%\hline
\midrule

\multirow{2}{2.25cm}{Multi-modal attention \cite{van2022gesture}} & JIGSAWS & \multirow{2}{2cm}{Kin + Vid} & \multirow{2}{2cm}{Suturing} & \multirow{2}{2cm}{Gestures} & \multirow{2}{2cm}{MA-TCN (Acausal)} & Accuracy: 86.8\% Edit: 91.4 \\
\cmidrule(lr){2-2} \cmidrule(lr){7-7} \hfill
 & own (dV) & & & & & Accuracy: 80.9\% Edit: 79.6\\
%\hline
\midrule

\multirow{3}{2cm}{Gesture Recognition Survey \cite{van2021gesture}} & \multirow{3}{2cm}{JIGSAWS} & Kinematics & \multirow{3}{2cm}{Suturing} & \multirow{3}{2cm}{Gestures} & MS-RNN \cite{farha2019ms} & Acc: 90.2\% Edit Score*: 89.5 \\
\cmidrule(lr){3-3} \cmidrule(lr){6-7} \hfill
 & & Video & & & Symm dilation + attention \cite{zhang2020symmetric} & Acc: 90.1\% Edit Score: 89.9\%  \\ % F1@10: 92.5\%\\
 \cmidrule(lr){3-3} \cmidrule(lr){6-7} \hfill
  & & Kin + Vid & & & Fusion-KV \cite{qin2020temporal} & Acc: 86.3\% Edit Score: 87.2 \\
%\hline
\midrule

\multirow{3}{2cm}{\textcolor{black}{PETRAW Challenge 2021 \cite{huaulme2022peg}}} & \multirow{3}{2cm}{\textcolor{black}{PETRAW}} & \textcolor{black}{Video} & \multirow{3}{2cm}{\textcolor{black}{Peg Transfer}} & \multirow{3}{2cm}{\textcolor{black}{Phases, Steps, and Activities}} & \textcolor{black}{SK} & \textcolor{black}{AD-Accuracy: 90.8\%} \\
\cmidrule(lr){3-3} \cmidrule(lr){6-7} \hfill
 & & \textcolor{black}{Kinematic} & & & \textcolor{black}{MedAIR} & \textcolor{black}{AD-Accuracy: 90.7\%} \\
 \cmidrule(lr){3-3} \cmidrule(lr){6-7} \hfill
 & & \textcolor{black}{Segmentation} & & & \textcolor{black}{SK} & \textcolor{black}{AD-Accuracy: 88.5\%} \\
 \cmidrule(lr){3-3} \cmidrule(lr){6-7} \hfill
 & & \textcolor{black}{Vid + Kin} & & & \textcolor{black}{NCC NEXT} & \textcolor{black}{AD-Accuracy: 93.1\%} \\
 \cmidrule(lr){3-3} \cmidrule(lr){6-7} \hfill
 & & \textcolor{black}{Vid+Kin+Seg} & & & \textcolor{black}{NCC NEXT} & \textcolor{black}{AD-Accuracy: 93.1\%} \\
\midrule

\multirow{4}{2cm}{Sim2Real Gesture Classification \cite{madapana2019desk, gonzalez2021dexterous}} & \multirow{4}{2cm}{DESK} & \multirow{4}{2cm}{Kinematics} & \multirow{4}{2cm}{Peg Transfer} & \multirow{4}{2cm}{Gestures} & \multirow{4}{2cm}{RF} & Simulator Acc: 86\% \\
%\cmidrule(lr){2-2} 
\cmidrule(lr){7-7} \hfill
 & & &  &  &  & Robot Acc: 95\% \\
%\cmidrule(lr){2-2} 
\cmidrule(lr){7-7} \hfill
 & &  & & &  & Sim2Real (0\% Real) Acc: 34\% \\
% \cmidrule(lr){2-2} 
\cmidrule(lr){7-7} \hfill
 &  & & & &  & Sim2Real (18\% Real) Acc: 85\% \\
% \cmidrule(lr){2-2} \cmidrule(lr){7-7} \hfill
%  & dVRK & & & & & Acc: 90\% \\
%\hline
\midrule

CholecTriplet2021 Challenge \cite{nwoye2023cholectriplet2021}
& \begin{tabular}[t]{@{} p{2cm}} CholecT50 \\ \end{tabular} & \begin{tabular}[t]{@{} p{2cm}}  Video  \end{tabular} &  \begin{tabular}[t]{@{} p{2.25cm}}  Laparoscopic cholecystectomy \end{tabular} & Action Triplets & Trequartista & \begin{tabular}[t]{@{} p{2cm}} $ AP_V $: 52.9 \\ $ AP_{IVT} $: 38.1 \end{tabular} \\
% \hline

\bottomrule
\end{tabular}
}
% \end{minipage}
\end{center}
\vspace{-0.5em}
\textcolor{black}{{\footnotesize * Normalized by maximum number of segments in any ground-truth sequence.}}
\vspace{-2em}
\end{table*}

\section{\textcolor{black}{Related Work}}
\label{sect:related_work}
Surgical workflow segmentation has been examined in different datasets with different tasks and at different levels of granularity as summarized in Table \ref{tab:related_work}.

\textbf{Datasets and Tasks:}
Recent works in surgical activity recognition perform comparative evaluation of their models across different datasets. For example, \cite{valderrama2022towards} developed the TAPIR model and found that it performed better on the MISAW dataset \cite{huaulme2021micro} than their PSI-AVA dataset for phase and step recognition, but did not examine the reason for this. \cite{yasar2020real} evaluated an LSTM using LOSO cross validation on the JIGSAWS dataset and their own dataset of Block Transfer on the RAVEN II. The LSTM achieved a higher accuracy for the Block Transfer task since it was a simpler task with a larger amount of data compared to the JIGSAWS tasks. 
\cite{li2022runtime} found that combining data from the Suturing and Needle Passing tasks in JIGSAWS could improve error detection performance because the gestures were kinematically similar. \textcolor{black}{\cite{itzkovich2021generalization} found that gesture recognition models trained on the JIGSAWS dataset did not generalize well to other dry-lab or clinical data.
Whereas previous works did not combine data from multiple datasets or tasks since the label definitions differed, in this paper we examine such aggregation in training surgical activity recognition models.}
% However, these works did not combine data from multiple datasets or tasks since the label definitions differed. %This suggests that there is a benefit to aggregating data for training.% models.

\textbf{Label Granularities:}
Surgical workflow \textcolor{black}{recognition has been examined at different levels of granularity as listed in the fifth column of Table \ref{tab:related_work}. Note that there are inconsistencies in label and granularity definitions across datasets. For example, the \textit{tasks} of Suturing, Knot Tying, and Peg Transfer in JIGSAWS and DESK are considered \textit{phases} in MISAW \cite{huaulme2021micro} and PETRAW \cite{huaulme2022peg}.} % is modeled at different levels with phases, steps, gestures, and actions. % in order of decreasing granularity. 
\cite{dipietro2019segmenting} trained a GRU for gesture and maneuver recognition on the JIGSAWS and MISTIC-SL datasets, respectively. Although the different datasets had different labels, the lower-level gesture recognition model had a higher error rate. The MISAW challenge \cite{huaulme2021micro} and HeiChole benchmark \cite{wagner2023comparative} datasets were labeled at multiple levels as well as the PSI-AVA dataset \cite{valderrama2022towards}. The best performing models from these works all showed decreasing performance metrics for finer-grained labels which highlights a significant challenge for fine-grained recognition. Interestingly, \cite{huaulme2021micro} found that multi-granularity recognition models performed better because such models may be learning that certain activities only occur during specific phases and steps. \textcolor{black}{Also, recent works on action triplet recognition in laparoscopic procedures focus on concurrent phase, step, and action recognition \cite{huaulme2022peg}. The poor performance of activity recognition models is a barrier to clinical applications, but understanding the relationship between granularity levels can address this challenge and guide model development. This work closes a gap between the gesture and action levels of the hierarchy by evaluating and comparing the performance of an activity recognition model at those granularities.}

\begin{comment}
\textbf{Robotic Systems:}
\cite{gonzalez2021dexterous} and \cite{madapana2019desk} examined gesture classification using transfer learning for the peg transfer task. 
% Their RF model was more accurate for trials performed on the Taurus robot compared to the dVRK, but the Taurus simulator had the lowest accuracy.
Complete transfer from the Taurus simulator to robot (0\% Taurus robot data in the training set) resulted in poor performance, but including a percentage of Taurus robot data in the training set allowed the model to achieve comparable performance to the Taurus simulator model. % on data from the Taurus robot.
%Although the performance of the model for complete transfer from the Taurus simulator to the Taurus robot (0\% Taurus robot data in the training set) resulted in poor performance, including a percentage of Taurus robot data in the training set allowed the model to achieve comparable performance to the Taurus simulator model on data from the Taurus robot. 
They found that models can transfer from simulated to real robots, or between real robots, although the latter required a greater percentage of data from the target domain to achieve comparable performance. Thus, future work may be able to leverage transfer learning across robotic systems and/or tasks to improve model performance.
\end{comment}

\section{Methods}
\begin{comment}
\subsection{Surgical Activity Recognition}
%\subsection{Gesture and Motion Primitive Recognition}
Surgical gesture recognition is a critical component of surgical process modeling \cite{lalys2014surgical,neumuth2011modeling}, skill assessment \cite{forestier2012classification, wang2021towards}, error detection \cite{yasar2020real,hutchinson2022analysis,li2022runtime}, and autonomy \cite{de2021first}. Much work has been done in developing models for gesture recognition mostly using data from the Suturing task in the JIGSAWS dataset as summarized in \cite{van2021gesture}. 
Previous works \cite{yasar2020real} and \cite{van2021gesture} have provided baseline gesture recognition model accuracies for various methods including LSTM \cite{yasar2020real}, TCN \cite{lea2016temporal}, SDSDL \cite{sefati2015learning}, and SC-CRF \cite{lea2015improved} that were trained and evaluated on the JIGSAWS dataset using only patient-side kinematic data. The SDSDL and SC-CRF models provide baseline accuracies for state-of-the-art models as shown in Table \ref{tab:LOUO}. However, existing gesture recognition models have been task-specific and restricted to specific datasets and gesture definitions.     
\end{comment}

This section presents our methods for the construction and evaluation of gesture and MP recognition models. %for the gesture and motion primitive levels.

\subsection{Data Pre-processing}
%As in \cite{lea2016temporal}, the kernel size is set to the average duration of the shortest action class (e.g., gesture or MP), the stride is 1, and the three layers have 32, 64, and 96 filters respectively. The loss function is cross-entropy and the model was trained using the Adam optimizer \cite{kingma2014adam}. 
The input to the activity recognition model is the time-series kinematic data, $x_t$, and the output is a transcript of class labels, $y_t$, one for each time-series sample, where each class label is selected from the finite set of gestures or MPs.
%For different combinations of kinematic variables as inputs to the activity recognition models w
\textcolor{black}{We experimented with different combinations of kinematic variables as inputs to the activity recognition models (while hyperparameter and cross validation settings were kept constant) and found that using only the position, linear velocity, and gripper angle kinematic variables resulted in the best performance.}
%Experimentation with different combinations of kinematic variables as inputs to the activity recognition models revealed that using only the position, linear velocity, and gripper angle resulted in the best performance. 
This is consistent with the best performing gesture recognition models that relied on kinematic data as reported in \cite{van2021gesture}.
The stride was 1, so there was no downsampling, and the kinematic data and gesture and MP labels were all at 30 Hz.

\subsection{Surgical Activity Recognition Model}
%\todo[inline]{one sentence explaining why we picked the TCN: speed, complexity, kinematic variables used...}
One of the fastest and best performing models that used only kinematic data for gesture recognition in \cite{van2021gesture} was the Temporal Convolutional Network (TCN). \textcolor{black}{The TCN is also used as a component in more complex state-of-the-art models such as MA-TCN \cite{van2022gesture} and MRG-Net \cite{long2021relational}. Thus, as a case study, we} adopt the TCN model from \cite{lea2016temporal} for activity recognition at both gesture and MP levels. This model has an encoder-decoder structure, each consisting of three convolutional layers with pooling, channel normalization, and upsampling. %as shown in Figure \todo[inline]{add TCN model figure}. 
As in \cite{lea2016temporal}, the kernel size is set to the average duration of the shortest activity class (e.g., gesture or MP), %the stride is 1, 
and the three layers have 32, 64, and 96 filters respectively. %The loss function is cross-entropy and the model was trained using the Adam optimizer \cite{kingma2014adam}. 
We used the cross-entropy loss function and Adam optimizer \cite{kingma2014adam}. %The input signal to the model is the time-series kinematic data, $x_t$, and the output is a class label, $y_t$, for each time-series sample. Since the stride was 1, there was no downsampling, so the kinematic data and gesture and MP labels were all at 30 Hz.
% %We trained and evaluated our models using kinematic data from the JIGSAWS dataset labeled with gestures and motion primitives (MPs). 

% Experimentation with different combinations of kinematic variables as inputs to the TCN revealed that using only the position, linear velocity, and gripper angle resulted in the best performance. This is consistent with the best performing gesture recognition models that relied on kinematic data as reported in \cite{van2021gesture}.

The learning rate and weight decay hyperparameters for all TCN models were selected based on a grid search of values by training on the JIGSAWS dataset with gesture labels for each cross validation setup. For LOUO models, the learning rate was 0.00005 and the weight decay was 0.0005. For LOTO models, the learning rate was 0.0001 and the weight decay was 0.001. These values were fixed for all models of their respective cross validation setup to analyze the effect of different training and label sets on model performance.
%so that the effect of different training and label sets on model performance could be analyzed.

We compare the performance of the TCN when trained with four different sets of labels: gestures, MPs for only the left side (Left MPs), MPs for only the right side (Right MPs), and MPs for both sides together (MPs).

%Recently...
%Our models predict just the verb of the MP using only kinematic data and could, thus, be compared to the results of action triplet recognition models \cite{nwoye2022rendezvous, li2022sirnet} that also predict just the verb but only based on video data. 
%\todo[inline]{ref table comparing results in results section}

\subsection{Model Generalization}
%\subsubsection{Cross Validation Setups}
We evaluate the generalization of the recognition models to unseen users/subjects and surgical tasks using two cross validation setups: Leave-One-User-Out (LOUO) from \cite{ahmidi2017dataset} and our novel Leave-One-Task-Out (LOTO).
%which assesses a model's ability to generalize to an unseen task by leveraging data from other tasks.

\subsubsection{Leave-One-User-Out (LOUO)}
LOUO is the standard cross validation setup for comparing gesture recognition models and is preferred over the Leave-One-Supertrial-Out (LOSO) method as it measures a model's ability to generalize to an unseen user \textcolor{black}{as} expected of a deployed model \cite{van2021gesture}. Since tasks from different aggregated datasets in COMPASS do not share the same subjects, we extended the LOUO setup from JIGSAWS \cite{ahmidi2017dataset} to include the new subjects, resulting in a maximum of 28 folds (corresponding to 28 users) when the model was trained on data from all tasks.

\subsubsection{Leave-One-Task-Out (LOTO)}
Existing datasets %with both kinematic and video data 
represent a limited number of trials, subjects, and tasks. This means that machine learning models trained on them will see subjects, trials, and \textit{tasks} that could be very different when they are deployed. In order to assess a model's ability to generalize to an unseen task, we introduce the Leave-One-Task-Out (LOTO) cross validation method. 

In the LOTO setup, all of the data for one \textcolor{black}{\textit{task}} was held out as the test set while the model was trained on %the rest of the tasks. 
all of the data for a set of other tasks. \textcolor{black}{Thus the model would be tested on all the trials of all subjects from an unseen task. For \textcolor{black}{an example fold,} a model could be trained on NP, KT, PT, PaS, and PoaP and tested on S.
This differs from the LOSO setup where a model would be tested on unseen \textit{trials} from a known subject of a known task.}
Similar to the existing LOSO and LOUO setups, average accuracy and edit score across the folds can be reported and used to compare models. However, examining each fold's performance and considering the relationship and similarity between the tasks in the training and test sets yields insights about the generalizability of the model to unseen tasks and the data needed to train a model.

\subsection{Task Combination for Training}
 The unified set of finer-grained MP labels enable combining data from different tasks across datasets which can improve the diversity and size of training data and model generalization. On the other hand, the gesture labels are specific to each dataset and only tasks with similar labels within that dataset can be combined. To evaluate the effect of label granularity on task generalization, we use data from different combinations of tasks in the aggregated datasets for model training in both LOUO and LOTO setups.
%For both the LOUO and LOTO setups, data from different tasks was grouped together if they shared the same context states, to improve the variability and size of training data. 
Using MPs, there were two combinations with similar context: S + NP = 'SNP' where both tasks have a task-specific needle state, and PT + PaS = 'PTPaS' where both tasks have a task-specific block state. Tasks could also be grouped together if they come from the same dataset: S + NP + KT = 'JIGSAWS' and PaS + PoaP = 'ROSMA'. Combining all of the data to train a model was referred to as 'All'. With gestures, only the SNP and JIGSAWS combinations could be used.
For LOTO, we also considered specific combinations of data that tested on one task but removed the contextually similar tasks (defined above) from the training set to assess the importance of augmenting the training set with data from similar tasks. % compared to tasks from the same dataset.
%Finally, for MPs we compared the performance of models trained on separate left and right robot arms vs. both sides combined. 

\subsection{Evaluation Metrics}
% We use the following standard metrics for the evaluation of gesture and MP recognition models.
\textcolor{black}{We use the standard metrics accuracy, edit score \cite{lea2016temporal}, and mean average precision (mAP) \cite{scikit-learn} for the evaluation of gesture and MP recognition models. Micro mAP is reported for each verb to account for class imbalance.}

\begin{comment}
\textbf{Accuracy:}
Given the lists of predicted and ground truth labels, the accuracy is the ratio of correctly classified samples divided by the total number of samples in a trial.
%The segmentation labels $G`$ and $P`$ represents labels in the form of $ {[A], [BBBB], [CC]}$. The accuracy calculates the ratio of the correctly classified samples over the total number of samples. 

\textbf{Edit Score:}
We report the edit score as defined in \cite{lea2016temporal} which uses the normalized Levenshtein edit distance, $ edit(G, P) $, by calculating the number of insertions, deletions, and replacements needed to transform the sequence of predicted labels $ P $  to match the ground truth sequence of labels $ G $. The edit score is normalized by the maximum length of the predicted and ground truth sequences as follows (with 100 representing the best prediction performance): %using Equation \ref{equn:edit} where 100 is the best and 0 is the worst.

\vspace{-0.5em}
\begin{equation}
    \text{Edit Score} = (1-\frac{edit(G, P)}{max(len(G), len(P))}) \times 100
    \label{equn:edit}
\end{equation}

\textbf{Mean Average Precision (mAP):}
Average Precision (AP) for a class is the average precision weighted by the increase in recall between thresholds calculated using Scikit-learn \cite{scikit-learn}. Then, the mean Average Precision (mAP) is the average AP for all classes.
\end{comment}

\section{Experimental Results}
Experiments were performed on a computer with an Intel Core i9 CPU @ 3.60GHz and 64GB RAM, running Linux Ubuntu 18.04 LTS, and an NVIDIA GeForce RTX 2070 GPU running CUDA 10.2, and the models were built and trained using Torch 1.10.1 \cite{paszke2019pytorch}. % to train the models.

%\section{Results}

\subsection{Gesture vs. Motion Primitive Recognition}
In this section we present the performance of TCN models in recognizing gestures and MPs in comparison to state-of-the-art models and with different combinations of data. %Using MP labels, different datasets can be combined and models trained on them can be compared. We show the limitations of using gestures when using a model to predict on an unseen task in the LOTO setup and how using MPs can help overcome this challenge. 
%Note that a direct performance comparison between the models for recognizing gestures vs. motion primitives is not reasonable since they are at different levels of the surgical hierarchy. 

%\subsubsection{LOUO}
\textcolor{black}{Tables \ref{tab:LOUO} and \ref{tab:LOUO_MPs}} compare the accuracies and edit scores averaged over the folds of the LOUO setup for the TCN models trained to recognize gestures \textcolor{black}{and MPs, respectively}. %Table \ref{tab:LOUO_MPs} shows the results for the MP models. 
Accuracies for two state-of-the-art models are also presented in Table \ref{tab:LOUO} against which our TCN model performs comparably or better. The TCN performed best on \textcolor{black}{S} alone achieving an accuracy of 84.6\% and an edit score of 87.7 which is also slightly better than the 79.6\% accuracy and 85.8 edit score reported by \cite{lea2016temporal} and comparable to the results of \cite{van2022gesture} for the TCN using only kinematic data (not shown in Table \ref{tab:LOUO}). %although they used kinematic data at 10 Hz while we used 30 Hz. 

\begin{table}[t!]
\vspace{0.5em}
\caption{Gesture recognition performance %TCN model accuracies, edit scores, and mAPs for gestures 
under the LOUO cross validation setup compared to state-of-the-art models using only kinematic data. Results for the state-of-the-art models were only available for the JIGSAWS tasks.}% from the JIGSAWS dataset.}
\centering
\label{tab:LOUO}
\resizebox{0.48\textwidth}{!}{
\begin{tabular}{cccc cc}
%\hline
\toprule
%\multicolumn{1}{c}{\begin{tabular}[c]{@{}c@{}} Tasks \\  \end{tabular}}  & \multicolumn{2}{c}{\begin{tabular}[c]{@{}c@{}}Gestures \\ (Acc/Edit Score)\end{tabular}}
Tasks & \multicolumn{3}{c}{Gestures} & \multicolumn{2}{c}{\textcolor{black}{Baselines}} \\
 & Acc (\%) & Edit Score & mAP & Acc (\%) & Model \\
% \midrule
\cmidrule(lr){1-1} \cmidrule(lr){2-4} \cmidrule(lr){5-6} 
%Tasks & \multicolumn{2}{l|}{Gestures} \\ \hline
PT & 73.5 & 83.8 & 80.7 &  &   \\ \cmidrule(lr){1-1} \cmidrule(lr){2-4} \cmidrule(lr){5-6}  %\cmidrule(lr){1-6}
S & 84.6 & 87.7 & 86.0 & \textcolor{black}{90.2} & \textcolor{black}{MS-RNN \cite{farha2019ms}} \\
NP & 78.4 & 85.2 & 86.4 & 75.3 & SC-CRF \cite{lea2015improved} \\
KT & 84.4 & 85.4 & 89.8 & 78.9 & SC-CRF \cite{lea2015improved} \\ \cmidrule(lr){1-1} \cmidrule(lr){2-4} \cmidrule(lr){5-6}  %\cmidrule(lr){1-6}
SNP & 81.4 & 85.2 & 85.1 &  &  \\
JIGSAWS & 80.9 & 82.0 & 85.7 &  &   \\
% PoaP & \multicolumn{1}{c}{} & \multicolumn{1}{c}{} \\
% PaS &\multicolumn{1}{c}{} & \multicolumn{1}{c}{} \\
% ROSMA &\multicolumn{1}{c}{} & \multicolumn{1}{c}{} \\

% PTPaS &\multicolumn{1}{c}{} & \multicolumn{1}{c}{} \\
% All &\multicolumn{1}{c}{} & \multicolumn{1}{c}{} \\ %\hline
\bottomrule
\end{tabular}
}
\vspace{-2em}
\end{table} 

\begin{table}[b!]
    \vspace{-1.5em}
    \centering
    \caption{%TCN model accuracies and edit scores for MPs
    MP recognition performance with different task combinations under the LOUO cross validation setup.}
    \label{tab:LOUO_MPs}
    \begin{tabular}{c cc cc cc}
    \toprule
    \multirow{2}{0.75cm}{Tasks} & \multicolumn{2}{c}{MPs} & \multicolumn{2}{c}{Left MPs} & \multicolumn{2}{c}{Right MPs} \\
     & Acc & Edit & Acc & Edit & Acc & Edit \\
    \midrule
    %\cmidrule(lr){1-1} \cmidrule(lr){2-3} \cmidrule(lr){4-5} \cmidrule(lr){6-7}
    S & 52.6 & 58.5 & \textbf{66.0} & \textbf{65.2} & 60.3 & 61.8 \\
    NP & 52.3 & 53.1 & \textbf{64.7} & \textbf{60.0} & 55.9 & 54.8 \\
    KT & 62.9 & 58.0 & \textbf{71.2} & \textbf{67.2} & 64.6 & 59.9 \\
    \cmidrule(lr){1-7}
    SNP & 55.2 & 56.2 & \textbf{66.5} & \textbf{62.2} & 59.5 & 61.1 \\
    JIGSAWS & 55.8 & 55.3 & \textbf{66.4} & \textbf{63.5} & 61.7 & 60.1 \\
    \cmidrule(lr){1-7}
    PoaP & 67.4 & 74.6 & \textbf{79.6} & 72.6 & 79.3 & \textbf{74.7} \\
    PaS & 70.2 & 76.5 & \textbf{80.0} & \textbf{77.6} & 78.5 & 75.9 \\
    \cmidrule(lr){1-7}
    ROSMA & 67.5 & \textbf{74.9} & \textbf{78.8} & 73.1 & 78.2 & 73.6 \\
    \cmidrule(lr){1-7}
    PT & 75.3 & 79.9 & 81.1 & 81.8 & \textbf{82.0} & \textbf{82.4} \\
    \cmidrule(lr){1-7}
    PTPaS & 70.3 & 76.4 & 78.5 & \textbf{77.8} & \textbf{78.8} & 77.4 \\
    \cmidrule(lr){1-7}
    All & 65.9 & 69.6 & \textbf{75.0} & 70.3 & 73.1 & \textbf{70.7} \\
    \bottomrule
    \end{tabular}
\end{table}

%We also observe that although the Suturing and Needle Passing tasks are contextually similar in the COMPASS framework and share the same gesture labels in the JIGSAWS dataset, the TCN only achieves accuracies and edit scores of 84.6\% and 87.7 for Suturing and 78.4\% and 85.2 for Needle Passing, respectively. 
%a 10\% decrease in accuracy in recognizing gestures in Needle Passing compared to Suturing. 
Despite KT only sharing two similar gestures and having a different task-specific context than the other two JIGSAWS tasks, the TCN's performance on KT is comparable to its performance on S (accuracy of 84.4\%, edit score of 85.4). When data from multiple tasks is combined for the 'SNP' and 'JIGSAWS' models, the TCN models' accuracies are only about the average of their performances on individual tasks while the edit score for the JIGSAWS model drops to 82.0 which is lower than any single task in that dataset. Thus, there does not appear to be much benefit to combining data from the JIGSAWS tasks at the gesture level. The PoaP and PaS tasks from the ROSMA dataset did not have gesture labels, so no gesture recognition models were trained for them. The PT task of the DESK dataset did have gesture labels although their definitions were much closer in scope to MPs rather than the more complex gestures of the JIGSAWS dataset. The TCN only achieves an accuracy of \textcolor{black}{73.5\%} for gesture recognition on the PT task which is comparably lower than the performance of any of the MP recognition models for this task in the LOUO setup \textcolor{black}{shown in Table \ref{tab:LOUO_MPs}}. For the JIGSAWS tasks, the gesture recognition models performed much better than MP recognition models (only considering verbs). This suggests that the definitions and granularity of the labels in the surgical hierarchy affect activity recognition performance. % so it is difficult to directly compare recognition models trained on labels of different definitions. 

\begin{comment}
\begin{figure}[b!]
    \centering
    %\vspace{-2em}
    \begin{minipage}{0.48\textwidth}
        \centering
        \includegraphics[%trim = 0.9in 1.6in 9in 0.7in, 
        width=\textwidth]{Figures/louo_acc.pdf}
        %\caption{first figure}
    \end{minipage}\hfill
    \begin{minipage}{0.48\textwidth}
        \centering
        \includegraphics[%viewport=180 10 320 540, clip=true, 
        width=\textwidth]{Figures/louo_edit.pdf}
    \end{minipage}\hfill
    % \begin{minipage}{0.48\textwidth}
    %     \centering
    %     \includegraphics[%viewport=180 10 320 540, clip=true, 
    %     width=\textwidth]{Figures/louo_mAP.pdf}
    % \end{minipage}
    \caption{TCN model accuracies and edit scores for MPs under the LOUO cross validation setup.}
    \label{fig:LOUO}
\end{figure}
\end{comment}

\begin{table*}[ht!]
\vspace{0.5em}
    \centering
    \caption{\textcolor{black}{Number of examples (\#) and mean average precision (mAP) of MPs for models trained on different combinations of tasks in the LOUO setup with micro mAP for all verbs (weighted by number of samples in each class).}}
    \label{tab:LOUO_mAPs_all}
    % \begin{tabular}{lp{1.5cm}c|cc}
    % \begin{tabular}{p{2cm} p{0.5cm} p{0.5cm} p{0.5cm} p{0.5cm} p{0.5cm} p{0.5cm} p{0.5cm} p{0.5cm} p{0.5cm} p{0.5cm} p{0.5cm} p{0.5cm} p{0.5cm} p{0.5cm}}
    \begin{tabular}{c rc rc rc rc rc rc rc}
        \toprule
        Tasks & \multicolumn{2}{c}{Grasp} & \multicolumn{2}{c}{Release} & \multicolumn{2}{c}{Touch} & \multicolumn{2}{c}{Untouch} & \multicolumn{2}{c}{Pull} & \multicolumn{2}{c}{Push} & \multicolumn{2}{c}{All verbs} \\
        \cmidrule(lr){2-3} \cmidrule(lr){4-5} \cmidrule(lr){6-7} \cmidrule(lr){8-9} \cmidrule(lr){10-11} \cmidrule(lr){12-13} \cmidrule(lr){14-15}
         & \# & mAP & \# & mAP & \# & mAP & \# & mAP & \# & mAP & \# & mAP & \# & mAP \\ 
        \midrule
        S & 471 & 57.6 & 441 & 48.7 & 518 & 58.1 & 314 & 27.6 & 194 & 72.2 & 179 & 55.1 & 2117 & 52.5\\ 
        NP & 373 & 63.0 & 365 & 57.0 & 330 & 57.0 & 206 & 16.2 & 114 & 69.1 & 119 & 34.2 & 1507 & 52.0 \\ 
        KT & 283 & 64.5 & 247 & 69.1 & 135 & 43.8 & 111 & 18.6 & 235 & 85.3 & 0 & N/A & 1011 & 62.7 \\
        \cmidrule(lr){1-15}
        SNP & 844 & 61.3 & 806 & 54.8 & 848 & 58.0 & 520 & 21.2 & 308 & 70.0 & 298 & 47.3 & 3624 & 52.9 \\
        JIGSAWS & 1127 & 62.2 & 1053 & 58.7 & 983 & 53.0 & 631 & 20.7 & 543 & 72.6 & 298 & 41.5 & 4635 & 53.7 \\
        \cmidrule(lr){1-15}
        PoaP & 577 & 52.8 & 556 & 55.3 & 1782 & 88.0 & 1261 & 47.2 & 525 & 58.3 & 2 & 33.5 & 4703 & 65.5 \\
        PaS & 824 & 50.2 & 776 & 50.3 & 1598 & 88.9 & 1131 & 45.7 & 0 & N/A & 0 & N/A & 4329 & 63.3 \\
        \cmidrule(lr){1-15}
        ROSMA & 1401 & 50.7 & 1332 & 53.1 & 3380 & 89.2 & 2392 & 45.3 & 525 & 59.2 & 2 & 5.1 & 9032 & 64.5 \\
        \cmidrule(lr){1-15}
        PT & 323 & 48.3 & 313 & 61.1 & 539 & 90.3 & 364 & 68.3 & 0 & N/A & 0 & N/A & 1539 & 70.3 \\ 
        \cmidrule(lr){1-15}
        PTPaS & 1147 & 48.7 & 1089 & 54.6 & 2137 & 89.8 & 1495 & 53.0 & 0 & N/A & 0 & N/A & 5868 & 65.9 \\
        \midrule
        All & 2851 & 54.5 & 2698 & 55.4 & 4902 & 79.5 & 3387 & 43.5 & 1068 & 65.7 & 300 & 37.7 & 15206 & 60.7 \\ 
        \bottomrule
    \end{tabular}   
    \vspace{-1.5em}
\end{table*}

By examining Table \ref{tab:LOUO_MPs}, we note that MP recognition performance is better for \textcolor{black}{the task in the DESK} dataset, and to a somewhat lesser extent for \textcolor{black}{tasks in the ROSMA} dataset, than for tasks in the JIGSAWS dataset. This could be because the JIGSAWS tasks (S, NP, KT) are more challenging with more complex grammar graphs~\cite{ahmidi2017dataset}, while the tasks in the ROSMA and DESK datasets are variations of a pick and place task with simpler grammar graphs. This is supported by the higher edit scores for the models trained on the ROSMA and DESK datasets than the models on the JIGSAWS dataset. Combining data at the MP level also resulted in performance metrics that are about the average of the individual tasks that were combined. But, training separate models for each side of the robot resulted in higher accuracies with comparable or better edit scores. % as seen in Fig \ref{fig: LOUO}. 
So, having separate annotations and models for the left and right arms of the robot can improve MP recognition performance. 

\begin{comment}
\begin{table}[t!]
    \vspace{0.5em}
    \centering
    \caption{Comparison of mean average precision (mAP) for recognizing MPs (using kinematic only) and corresponding verbs in state-of-the-art action triplet models (using video only) in the LOUO setup.}
    \label{tab:LOUO_mAPs}
    \begin{tabular}{lp{1.5cm}c|cc}
        \toprule
        Model & Tasks & All verbs & Grasp & Pull/Retract \\
        \midrule
        \multirow{11}{*}{TCN} & S &  53.2  & 57.6 & 72.2 \\
        & NP &  49.4 & 63.0 & 69.1 \\
        & KT & 56.3 & \textbf{64.5} & \textbf{85.3} \\
        \cmidrule(lr){2-5}
        & SNP & 53.1 & 61.3 & 70.0 \\
        & JIGSAWS & 51.5 & 62.2 & 72.6 \\
        \cmidrule(lr){2-5}
        & PoaP & 59.4 & 52.8 & 58.3 \\
        & PaS & 58.8 & 50.2 & N/A \\
        \cmidrule(lr){2-5}
        & ROSMA & 57.9 & 50.7 & 59.2 \\
        \cmidrule(lr){2-5}
        & PT & \textbf{67.0} & 48.3 & N/A \\
        \cmidrule(lr){2-5}
        & PTPaS & 61.5 & 48.7 & N/A \\
        \cmidrule(lr){2-5}
        & All & 58.5 & 54.5 & 65.7 \\
        \midrule
        TCN \cite{nwoye2022rendezvous} & \multirow{3}{2cm}{Laparoscopic cholecystectomy} & 29.4 & 24.9 & 80.2 \\
        Tripnet \cite{nwoye2022rendezvous} &  & 54.5 & 45.8 & 88.1 \\
        Rendezvous \cite{nwoye2022rendezvous} &  & \textbf{60.7} & \textbf{60.4} & \textbf{90.5} \\  
        
        \bottomrule
    \end{tabular}
\vspace{-1em}    
\end{table}
\end{comment}

\textcolor{black}{Furthermore,} %MP recognition is similar to the recognition of the verbs of action triplets. 
\textcolor{black}{Table \ref{tab:LOUO_mAPs_all} shows the mAPs for each MP and micro average over all verbs for the MP recognition models in the LOUO set up. We note that class imbalance may have caused differences between the macro and micro mAPs for tasks from the DESK and ROSMA datasets where MPs with a greater number of instances sometimes had higher mAPs.}
%The closest fine-grained activity recognition models that can be used for assessing these results are verb recognition models in action triplets \cite{nwoye2022rendezvous}. %All of our TCN models using only kinematic data perform better than the baseline TCN model using only video data from \cite{nwoye2022rendezvous}, which achieves an average mAP of 29.4 for all verbs and mAP of 24.9 for "Grasp". %Some of our models, such as PT with an mAP of 70.3, outperform the state-of-the-art Rendezvous model \cite{nwoye2022rendezvous} which achieves an mAP of 60.7. 
%However, this only provides a sense of how well our models perform compared to the state-of-the-art verb recognition models. A direct comparison to action triplet models is not fair as the data (kinematic vs. video) and tasks (robotic dry-lab vs. real laparoscopic surgery) are different.}
\textcolor{black}{None} of these MP models perform as well as the gesture recognition models for the JIGSAWS tasks as listed in Table \ref{tab:LOUO}, which achieve mAPs of up to 89.8. So additional work is needed to improve fine-grained activity recognition performance.
\textcolor{black}{Although the recognition models of \cite{nwoye2022rendezvous} have been evaluated for verb recognition performance, a direct comparison to action triplet models is not fair as the data (kinematic vs. video) and tasks (robotic dry-lab vs. real laparoscopic surgery) are different.}

\begin{table*}[ht!]
    \vspace{0.5em}
    \centering
    \caption{%TCN model accuracies and edit scores for MPs and gestures for folds of the LOTO cross validation setup.
    MP and gesture recognition performance with different task combinations under LOTO cross validation setup.}
    \label{tab:LOTO_all}
    \resizebox{\textwidth}{!}{\begin{tabular}{*{1}{P{5.5mm}} *{6}{P{5.5mm}} *{8}{P{8mm}}} 
    \toprule
    %\hline
    %Test Set & \multicolumn{6}{P{5.5mm}}{Training Sets (Task combinations)} & \multicolumn{2}{P{8mm}}{{Gestures \ (Acc/Edit Score)}} & \multicolumn{2}{P{8mm}}{{MPs \ (Acc/Edit Score)}} & \multicolumn{2}{P{8mm}}{{Left MPs \ (Acc/Edit Score)}} & \multicolumn{2}{P{8mm}}{{Right MPs \ (Acc/Edit Score)}} \\ 
    %Test Set & \multicolumn{6}{c}{Training Sets \ (Task combinations)} & \multicolumn{2}{c}{{Gestures (Acc/Edit Score)}} & \multicolumn{2}{c}{{MPs \ (Acc/Edit Score)}} & \multicolumn{2}{c}{{Left MPs \ (Acc/Edit Score)}} & \multicolumn{2}{c}{{Right MPs \ (Acc/Edit Score)}} \\ 
    
    % \multicolumn{1}{c}{\begin{tabular}[c]{@{}c@{}} Test Set \\  \end{tabular}} & \multicolumn{6}{c}{\begin{tabular}[c]{@{}c@{}}Training Set\\ (Task combinations)\end{tabular}} & \multicolumn{2}{c}{\begin{tabular}[c]{@{}c@{}}Gestures \\ (Acc/Edit Score)\end{tabular}} & \multicolumn{2}{c}{\begin{tabular}[c]{@{}c@{}}MPs\\ (Acc/Edit Score)\end{tabular}} & \multicolumn{2}{c}{\begin{tabular}[c]{@{}c@{}}Left MPs\\ (Acc/Edit Score)\end{tabular}} & \multicolumn{2}{c}{\begin{tabular}[c]{@{}c@{}}Right MPs\\ (Acc/Edit Score)\end{tabular}}
    % \\
    
    \multicolumn{1}{c}{Test Set} & \multicolumn{6}{c}{Training Set} & \multicolumn{2}{c}{Gestures} & \multicolumn{2}{c}{MPs} & \multicolumn{2}{c}{Left MPs} & \multicolumn{2}{c}{Right MPs} \\
     & \multicolumn{6}{c}{(Task combinations)} & Acc & Edit & Acc & Edit & Acc & Edit & Acc & Edit
    %& \multicolumn{2}{c}{Gestures} & \multicolumn{2}{c}{MPs} & \multicolumn{2}{c}{Left MPs} & \multicolumn{2}{c}{Right MPs} \\ & \multicolumn{6}{c}{(Task combinations)} & \multicolumn{2}{c}{Acc/Edit Score} & \multicolumn{2}{c}{Acc/Edit Score} & \multicolumn{2}{c}{Acc/Edit Score} & \multicolumn{2}{c}{Acc/Edit Score} \\
    \\\cmidrule(lr){1-1}\cmidrule(lr){2-7}\cmidrule(lr){8-15}
    S &  & NP & KT & PT & PaS & PoaP &  &  & 39.0 & \textbf{49.0} & 62.3 & \textbf{59.4} & 42.9 & \textbf{58.5} \\
    S &  &  & KT & PT & PaS & PoaP &  &  & 25.3 & 40.2 & 41.3 & 50.2 & 34.8 & 42.5 \\
    S &  & NP & KT &  &  &  & 24.4 & 33.9 & 43.2 & 48.3 & 56.2 & 52.7 & 46.3 & 48.2 \\
    S &  & NP &  &  &  &  & \textbf{48.5} & \textbf{70.5} & \textbf{44.0} & 47.7 & \textbf{62.7} & 58.7 & \textbf{50.1} & 54.7 \\\midrule
    NP & S &  & KT & PT & PaS & PoaP &  &  & 40.8 & 48.5 & \textbf{54.1} & \textbf{55.9} & 41.6 & 46.4 \\
    NP &  &  & KT & PT & PaS & PoaP &  &  & 35.6 & 44.5 & 46.2 & 51.9 & 34.4 & 39.9 \\
    NP & S &  & KT &  &  &  & 28.8 & 38.2 & 37.2 & 46.9 & 49.9 & 52.8 & 44.7 & 48.3 \\
    NP & S &  &  &  &  &  & \textbf{37.9} & \textbf{52.7} & \textbf{42.2} & \textbf{48.6} & 52.2 & 51.9 & \textbf{46.0} & \textbf{52.8} \\\midrule
    KT & S & NP &  & PT & PaS & PoaP &  &  & \textbf{33.3} & 40.2 & 47.2 & \textbf{51.9} & \textbf{35.2} & 39.8 \\
    KT &  &  &  & PT & PaS & PoaP &  &  & 22.6 & 37.5 & 37.7 & 36.5 & 25.1 & 36.8 \\
    KT & S & NP &  &  &  &  & \textbf{6.8} & \textbf{9.3} & 29.7 & \textbf{40.5} & \textbf{48.1} & 50.4 & 34.5 & \textbf{42.8} \\\midrule
    PT & S & NP & KT &  & PaS & PoaP &  &  & \textbf{53.1} & \textbf{48.0} & \textbf{55.9} & \textbf{42.0} & 43.9 & 38.6 \\
    PT & S & NP & KT &  &  & PoaP &  &  & 44.5 & 44.4 & 49.0 & 37.6 & \textbf{55.3} & \textbf{44.8} \\
    PT &  &  &  &  & PaS &  &  &  & 48.0 & 37.6 & 51.1 & 40.3 & 52.6 & 43.5 \\\midrule 
    PaS & S & NP & KT & PT &  & PoaP &  &  & 58.1 & \textbf{65.5} & 58.8 & \textbf{60.5} & 61.1 & \textbf{58.0} \\
    PaS & S & NP & KT &  &  & PoaP &  &  & 60.7 & 65.0 & 58.5 & 58.5 & 61.4 & 57.7 \\
    PaS &  &  &  & PT &  & PoaP &  &  & 58.0 & 64.1 & \textbf{65.8} & 58.3 & \textbf{63.9} & 57.2 \\
    PaS &  &  &  & PT &  &  &  &  & \textbf{61.0} & 37.5 & 42.5 & 54.6 & 55.0 & 42.9 \\
    PaS &  &  &  &  &  & PoaP &  &  & 58.4 & 62.9 & 59.5 & 57.2 & 59.8 & 56.1 \\\midrule
    PoaP & S & NP & KT & PT & PaS &  &  &  & \textbf{56.5} & \textbf{64.2} & \textbf{59.1} & \textbf{50.7} & \textbf{58.5} & \textbf{49.8} \\
    PoaP & S & NP & KT & PT &  &  &  &  & 53.4 & 47.8 & 50.4 & 45.9 & 36.0 & 43.9 \\
    PoaP &  &  &  &  & PaS &  &  &  & 54.8 & 63.1 & 57.8 & 44.9 & 58.0 & 45.2 \\\bottomrule
    %\hline
\end{tabular}
}
\vspace{-1.5em}
\end{table*}

\subsection{Model Generalization}
%\subsubsection{LOTO}
% Although performance metrics have been reported as the average across all the folds in the LOSO or LOUO setups, examining the results of individual folds in the LOTO setup can yield important insights about the ability of the model to generalize to an unseen task. This is important to analyze since datasets available for training ML models are limited in size and variety, so a deployed model will likely see tasks not represented in its training data. 

Table \ref{tab:LOTO_all} reports the accuracies and edit scores for models trained with different combinations of data in the LOTO setup and immediately shows limitations of existing gesture definitions. Note that only the JIGSAWS dataset had gesture labels that could be used in the LOTO setup, so gesture recognition models using tasks from different datasets could not be trained because gesture labels were not present or were not compatible. 
We observe that splitting the MP labels into separate transcripts and training separate models for the left and right arms of the robot generally results in improved accuracies compared to having a single model.

We find that a gesture recognition model trained on S or NP is able to transfer to NP or S, respectively, but when KT is added to the training set, performance is severely decreased. Specifically, a model tested on S drops from an accuracy of 48.5\% to 24.4\%, and a model tested on NP drops from 37.9\% to 28.8\% when KT is added to the training set. This is due to the lack of generalizable gesture labels between these tasks since S and NP have an almost completely different set of gestures than KT. Thus, gesture recognition for the KT task using a model trained on S and NP is particularly poor with an accuracy of only 6.8\%. Hence, at the gesture level, combining data from different tasks is not beneficial for a model that must predict on an unseen task.

Comparatively, when MPs are used, the model is able to predict on a new task like KT by leveraging information learned from other tasks that are dissimilar to it such as S and NP. % when predicting on a new task like KT. 
%By adding data from a dissimilar task, models for S and NP perform almost as well with KT in their training sets as they do without.
Adding data from a dissimilar task has a much smaller detrimental effect at the MP level than at the gesture level. For example, the model's accuracy drops less than 1\% for S and 5\% for NP when KT is added to the training set. % Specific to L/R models: More importantly, the model performs much better on the unseen dissimilar task of KT when trained on only S and NP. %If the MP labels are split into left and right MP labels and separate models are trained for each side of the robot, 
%Again, accuracies are always better with MP labels than gesture labels for a given test and training set combination. 

When the model must predict MPs on a dissimilar task with a different task-specific context state, %, namely KT and PoaP, 
then combining data from all tasks results in better performance compared to using only data in the same dataset. KT improves from an accuracy of 29.7\% to 33.\textcolor{black}{3}\% and PoaP improves from 5\textcolor{black}{4.8}\% to 56.\textcolor{black}{5}\% by including data from other datasets.

%We find that combining similar tasks based on having the same task-specific context state, namely S with NP and PT with PaS, usually results in models with higher accuracies than combining data from all tasks together. The exception is PaS where the Left MP model has the best performance when PT is not included in the training set and the Right MP model is trained on only PT and PoaP. 
%Notably, for PaS, combining both a similar task, PT, and a task from the same original dataset, PoaP, results in the best performance for the Right MP model which suggests that there may be a benefit to including data from the same subject even if the task is dissimilar. Likewise, the model also performs best when trained on a task that is both similar and from the same dataset, specifically training on S and testing on NP and vice versa. This means that including a similar task by the same subject or subjects enables the model to generalize better to an unseen task. 
%With separate models for the left and right sides, if a similar task is not present, as for KT and PoaP, then including only data from the same original dataset does best. This suggests that having data from the same subjects but different tasks is more important than data from different tasks and different subjects in helping the model generalize to an unseen task. % helps the model generalize to an unseen task.

For S and NP, we observe that models trained with data from the same dataset and with the same task-specific state variable perform better than models including data from the same dataset but without the same task-specific state variable. However, the opposite is true for PaS where models whose training sets included PoaP (same dataset) but not PT (same task-specific state variable) sometimes performed better.

For KT and PoaP, even though data with the same context was not available, models whose training sets included tasks from the same dataset generally performed better than models whose training sets did not. The poorest performing models for PaS were trained with data that only included PT, %but not data from the same dataset. 
even though they had the same task-specific state variables.
For PT, some models that included PaS (same task-specific state variable) performed better than those that did not.
Since tasks from the same dataset were performed by the same subjects, models whose training sets included tasks from the same dataset are tested on different tasks performed by known subjects. This is somewhat similar to the Leave-One-Supertrial Out (LOSO) cross validation method where models are tested on unseen trials performed by known subjects. Models evaluated using the LOSO method perform better than those using the LOUO method which suggests that including data from the same subjects may improve model performance. However, additional data and tests would be needed to determine if it is this or another feature of the dataset that is responsible for the performance improvement. \textcolor{black}{Additional evaluations are also needed to verify that MPs enable task generalization for other types of models such as transformers \cite{shi2022recognition}.}

\section{Discussion and Conclusion}

%In summary, we present a framework for modeling surgical tasks as motion primitives that cause changes in surgical context and apply it to three publicly available datasets to create an aggregate dataset of kinematic and video data along with context and motion primitive labels. %Our method for labeling context achieves substantial to near-perfect agreement between annotators and expert surgeons. %Additionally, we show that existing gesture definitions limit the combination of task data from different datasets. 

%We find that gesture recognition models outperform motion primitive recognition models in the LOUO setup, but higher motion primitive recognition accuracy can be achieved by training separate models for the left and right arms of the robot. 

% Something here.. in summary...
In summary, we compare the performance of activity recognition in a case study of TCN models at different levels of the surgical hierarchy, evaluate their generalizability to unseen users and tasks, and draw insights from the combinations of tasks used to train these models.

We find that gesture-level recognition models perform better than motion primitive-level recognition models under the LOUO cross validation method which is consistent with the observations of \cite{huaulme2021micro}. %Our models can also achieve higher mAPs \textcolor{black}{in verb recognition than some of the video-only verb recognition models for laparoscopic tasks~\cite{nwoye2022rendezvous}}. %TCN %, Tripnet, 
%and Rendezvous models from \cite{nwoye2022rendezvous}. 
%\textcolor{black}{But a direct comparison between kinematic-based models for dry-lab vs. video-based models for real procedures is not possible.} % which supports the use of kinematic data for activity recognition models. 
Our models achieve \textcolor{black}{comparable or better accuracies} %the same or slightly better performance (average accuracy and mean average precision (mAP)) than 
\textcolor{black}{than} state-of-the-art in recognizing gestures (from JIGSAWS). % and similar actions (Grasp and Pull/Retract).

% We also introduce the Leave-One-Task-Out (LOTO) cross validation setup, and perform the first evaluation of a surgical activity recognition model in terms of its ability to generalize to an unseen task.
Using motion primitives, we combine data from different datasets, tasks, and subjects and find that having separate models for the left and right sides improves performance. 
We also introduce the Leave-One-Task-Out (LOTO) cross validation setup, and perform the first evaluation of a surgical activity recognition model in terms of its ability to generalize to an unseen task.
\textcolor{black}{When tested on a task from a specific dataset, the model performed better if data from other tasks in that dataset were included in training.} %model tested on a specific task performed better if the training data included other tasks from the same original dataset.} %Specifically, we find that models can perform better for an unseen task if trained on data that includes tasks in the same original dataset.} 
Also, models for tasks with different task-specific state variables perform best when data for all other tasks is aggregated for their training. Similarly, \cite{rahman2019transferring} evaluated the performance of surgeme classification models in sim2real domain transfer using different data percentages in the target domain and found that this improved the accuracies of their models. Thus, improved performance may be achieved by including a small percentage of data from the target test task in the training dataset.

\textcolor{black}{Future work will focus on evaluating the task generalization of other state-of-the-art recognition models (e.g., recurrent neural networks and transformers) using both kinematic and vision data as well as other tasks and datasets.}

\section*{Acknowledgment}
This work was supported in part by the National Science Foundation grants DGE-1842490, DGE-1829004, and CNS-2146295 and by the Engineering-in-Medicine center at the University of Virginia. We thank Dr. Schenkman, Dr. Cantrell, and Dr. Chen for their medical feedback.

% \vspace{-0.5em}
\section*{Competing Interests}
The authors have no competing interests to declare that are relevant to the content of this article.
\vspace{-0.25em}

\bibliographystyle{IEEEtran}
\bibliography{main.bib}

% \bibliographystyle{plain}
% \bibliography{main.bib}
%\bibliography{sn-bibliography}% common bib file
%% if required, the content of .bbl file can be included here once bbl is generated
%%\input sn-article.bbl

%% Default %%
%%\input sn-sample-bib.tex%

\end{document}